%% file: main.tex
\documentclass[9pt]{article}

\usepackage[a4paper, margin=1in]{geometry}
\usepackage[utf8]{inputenc}
\usepackage[T1]{fontenc}
\usepackage{hyperref}
\usepackage{url}
\usepackage{booktabs}
\usepackage{tabularx}
\usepackage{longtable}
\usepackage{amsfonts}
\usepackage{amsmath}
\usepackage{microtype}
\usepackage{xcolor}
\usepackage{graphicx}
\usepackage[numbers,sort&compress]{natbib}
\usepackage{authblk}
\usepackage{subcaption}

\hypersetup{colorlinks=true, allcolors=blue!50!black}

\makeatletter
\renewcommand\paragraph{\@startsection{paragraph}{4}{\z@}%
  {1.2ex \@plus .4ex \@minus .2ex}{-1em}{\normalfont\normalsize\bfseries}}
\makeatother

\newcommand{\dataset}{Open-fMIND}

\input{released_cites}

\newcommand{\nDatasets}{1{,}066}        %
\newcommand{\nSubjects}{45{,}067}       %
\newcommand{\rawTB}{67.0}               %
\newcommand{\nTasks}{1{,}913}           %
\newcommand{\medSubjects}{25}           %

\newcommand{\nKnownDatasets}{387}       %
\newcommand{\nExcluded}{27}             %
\newcommand{\nExclNoBold}{24}           %
\newcommand{\nExclDeriv}{2}             %
\newcommand{\nExclManual}{1}            %
\newcommand{\nDlFailed}{1}              %
\newcommand{\nDownloaded}{359}          %
\newcommand{\relDatasets}{322}          %
\newcommand{\nKnownSubjects}{14{,}016}  %
\newcommand{\effSubjects}{11{,}379}     %
\newcommand{\nRawRuns}{52{,}505}        %
\newcommand{\relRuns}{45{,}731}         %
\newcommand{\nShortRuns}{712}           %
\newcommand{\prepDatasets}{320}         %
\newcommand{\prepSubjects}{11{,}320}    %
\newcommand{\prepRuns}{45{,}019}        %
\newcommand{\nAtlases}{52}              %
\newcommand{\nFamilies}{14}             %
\newcommand{\relMatrices}{2.4}          %
\newcommand{\relSeriesQ}{2.38}          %
\newcommand{\relSeriesP}{2.34}          %
\newcommand{\releaseTB}{4.32}           %
\newcommand{\nShards}{1{,}787}          %
\newcommand{\relGB}{4{,}321}              %
\newcommand{\relTasks}{429}             %
\newcommand{\relMultiSession}{63}       %

\newcommand{\nDemogDatasets}{263}       %
\newcommand{\nDemogSubjects}{9{,}428}   %
\newcommand{\pctSexKnown}{69.4}         %
\newcommand{\pctFemale}{38.3}
\newcommand{\pctMale}{31.0}
\newcommand{\pctOtherSex}{0.1}
\newcommand{\pctAgeKnown}{71.7}         %
\newcommand{\meanAge}{28.7}
\newcommand{\medianAge}{24.0}

\title{\textbf{\dataset: A Large-Scale Open-Access fMRI Corpus
for Pretraining Brain Foundation Models}}

\author[1]{Carlo Alberto Barbano\thanks{Corresponding author: \texttt{carlo.barbano@inria.fr}}}
\author[1]{Giovanni Marraffini}
\author[1]{Demian Wassermann}

\affil[1]{MIND, Université Paris-Saclay, Inria, CEA, Palaiseau, 91120, France
}
\date{}

\begin{document}

\maketitle

\begin{abstract}
\noindent
Self-supervised pretraining has transformed language and vision, but its
application to functional MRI (fMRI) is constrained by data access. The largest
standardized fMRI resources are governed by data use agreements and dominated
by a small number of acquisition protocols, while the thousands of openly
licensed datasets on OpenNeuro are individually small, heterogeneously
organized, and unprocessed. We present the first release of \dataset{}, a
uniformly preprocessed, openly accessible corpus of parcellated fMRI timeseries
from \relDatasets{} OpenNeuro studies, comprising \effSubjects{} subjects and
\relRuns{} functional runs. Each run is preprocessed with fMRIPrep and
parcellated under \nAtlases{} atlases spanning \nFamilies{} families, yielding a
massive \emph{\relMatrices{}~million timeseries}.
Timeseries are released either denoised with our default pipeline
(\texttt{preproc}) or as the fMRIPrep output with its confound files
(\texttt{quasiraw}), with functional connectivity matrices, per-run
provenance and motion metadata.
\dataset{} is intended as a pretraining resource for self-supervised brain
representation learning. Pretraining on it yields downstream performance comparable to that
obtained with UK Biobank.
\end{abstract}

\section{Background \& Summary}
\label{sec:background}

\begin{figure}[t]
    \centering
    \includegraphics[width=\linewidth]{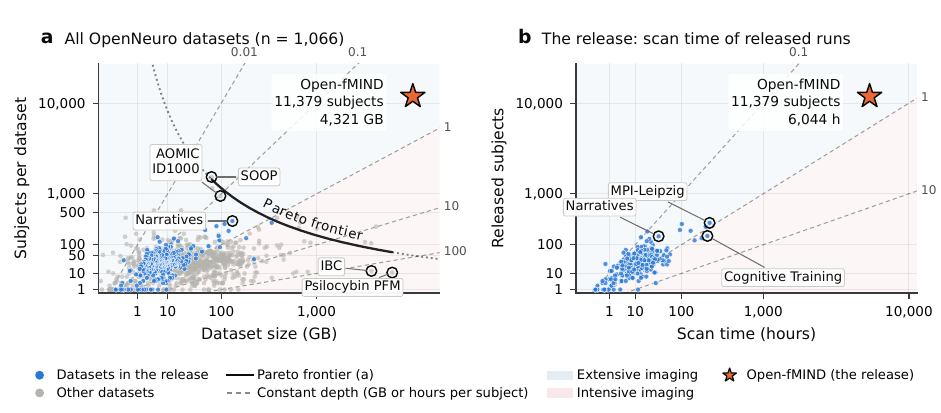}
    \caption{\textbf{\dataset{} pools many small datasets into a corpus larger than any single one of them.} (a) The \nDatasets{} OpenNeuro MRI datasets, by size and number of subjects. Blue points are datasets with released runs in \dataset{}, grey points the others. Dashed lines mark constant size per subject, and the black line is the Pareto frontier (no dataset has both more subjects and more data). The star is the release itself: its \effSubjects{} subjects and the size of its files (\relGB~GB, both versions and all atlases). %
    (b) Scan time against subjects for the released datasets, restricted to released subjects and runs. %
    }
    \label{fig:scale}
    \vspace{1.5em}
    \captionof{table}{Comparison with existing resources. Subject counts for HCP are from the
    S1200 release, for UK Biobank are participants with fMRI, for ABCD are from Release 5.1, and for IBC are the participants of the cumulative dataset up to the fifth release.}
    \label{tab:resources}
    \small
    \begin{tabular}{lccccc}
      \toprule
      Resource & Access & Subjects & Studies/sites & Task diversity & Release \\
      \midrule
      HCP \citep{hcp}        & Gated & 1{,}206        & 1     & Low-Med         & Raw + deriv. \\
      UK Biobank \citep{ukbiobank} & Gated & $>$40{,}000 & 1 & Low (rest+task) & Raw + deriv. \\
      ABCD \citep{abcd}       & Gated & 11{,}868       & 21    & Low-Med         & Raw + deriv. \\
      IBC \citep{ibc5}       & Open  & 13             & 1     & High (67 tasks) & Raw + deriv. \\
      OpenNeuro (single) & Open  & $\sim$\medSubjects~(median) & 1 & varies  & Raw \\
      \textbf{\dataset{}} & \textbf{Open} & \textbf{\effSubjects} &
        \textbf{\relDatasets} & \textbf{High (\relTasks~labels)} &
        \textbf{Derivatives} \\
      \bottomrule
    \end{tabular}
\end{figure}

Self-supervised pretraining on large, diverse corpora has driven foundation
models across language and vision, and a growing body of work brings the same
recipe to functional brain imaging \citep{brainlm, brainjepa}. The central
obstacle is data. The largest standardized fMRI resources, the Human Connectome
Project (HCP)~\citep{hcp}, UK Biobank \citep{ukbiobank}, and ABCD
\citep{abcd}, are large and well curated, but access is restricted by data use
agreements, and the data are dominated by a few acquisition protocols, mainly
resting state and a fixed task battery. OpenNeuro \citep{openneuro} hosts
thousands of open datasets in BIDS format \citep{bids}, but each is small,
heterogeneously organized, and either unprocessed or processed in a different way. Turning this long tail into a
pretraining corpus requires retrieving, preprocessing, and denoising hundreds of
studies in a uniform way, and this effort has kept the data largely
unused at scale.

We present the first release of \dataset{}: \relMatrices{}~million parcellated
region-of-interest (ROI) timeseries for \effSubjects{} subjects and \relRuns{} functional runs from
\relDatasets{} %
fMRI datasets on OpenNeuro (listed in Table~\ref{tab:released-datasets}, Appendix~\ref{app:datasets}).
Later releases will
extend the corpus to the full set. Individual datasets are small (the median
OpenNeuro fMRI dataset has \medSubjects{} subjects), whereas the pooled release
holds more subjects and more data than any single one of them
(Fig.~\ref{fig:scale}). Each run is provided in two versions, the
parcellated signal without denoising (\texttt{quasiraw}) and a denoised version
(\texttt{preproc}), under \nAtlases{} atlases spanning \nFamilies{} parcellation
families. At \effSubjects{} subjects, the corpus is comparable in subject count
to ABCD and requires no data use agreement, and its datasets carry \relTasks{}
distinct task labels, far more than rest-dominated resources. To our knowledge, \dataset{} is the largest fMRI
corpus that is simultaneously open, preprocessed, and model-ready
(Table~\ref{tab:resources}).
Two design choices shape the release. First, every run is parcellated under all
atlases, because hard parcellations (Schaefer, AAL), soft, spatially overlapping
decompositions (DiFuMo, Smith), and functionally and anatomically derived atlases
make different assumptions about what a region is. The atlas axis supports
studies of parcellation sensitivity and representation learning that treats the
parcellation as a view or augmentation (Usage Notes). Second, we release no
phenotypic targets, diagnostic labels, or behavioral scores. Phenotypic metadata
on OpenNeuro follows no enforced standard, so harmonizing it across
independently designed studies is a large effort and still an open question
\citep[e.g.,][]{neurobagel}. Self-supervised pretraining does not need labels,
so we limit the release to imaging data that is processed uniformly and leave
label curation to downstream users. The data-quality analyses in Technical
Validation are therefore label-free.

\section{Methods}
\label{sec:methods}

\dataset{} is produced by a pipeline with two parts: acquisition of the input
data, and preprocessing with parcellation and denoising.

\subsection{Input data and acquisition}

We query the OpenNeuro GraphQL API for datasets with functional MRI in human
participants and record, for each, the accession ID, title, modalities, subject
and session counts, task labels, file counts, size, species, and diagnostic
status where declared. The query returns \nDatasets{} datasets,
\nSubjects{} subjects, \rawTB{}~TB of raw data, and \nTasks{} unique task
labels, and defines the scope of the openly available fMRI data on OpenNeuro.
This first release covers a subset of these datasets, and future releases will
extend the corpus to the full set. The datasets are retrieved as BIDS directories and
inspected before preprocessing. Of the \nKnownDatasets{} datasets processed for
this release, \nExcluded{} are removed at inspection (\nExclNoBold{} without BOLD
runs, \nExclDeriv{} derivative datasets, and \nExclManual{} explicitly excluded)
and \nDlFailed{} fails to download, which leaves \nDownloaded{} datasets
(Table~\ref{tab:funnel}).

\subsection{Preprocessing}

The pipeline has four steps: preprocessing with fMRIPrep, parcellation under all
atlases, denoising, and computation of connectivity matrices. We release the
result in two versions. The first, \texttt{quasiraw}, is the output of fMRIPrep
followed by parcellation, with no denoising. The fMRIPrep confounds file of each
run is released next to its timeseries, so that users can apply their own
denoising pipeline to \texttt{quasiraw}. The second, \texttt{preproc}, adds a
default denoising step designed to be broadly acceptable for both resting-state
and task data. Each version has one timeseries for each run and atlas, and both
include connectivity matrices.

\paragraph{Step 1: fMRIPrep.} fMRIPrep \citep{fmriprep} corrects head motion,
aligns each functional run with the anatomical image of the subject, and
resamples the data to the \texttt{MNI152NLin2009cAsym} template at 2~mm
isotropic resolution in a single interpolation step. Slice-timing and
susceptibility distortion corrections are applied when the dataset provides the
information they need. For each run, fMRIPrep also writes a table of confound
signals (motion parameters, framewise displacement, and mean white matter and
cerebrospinal fluid signals, among others), which denoising uses. Each subject
is processed as a separate cluster job from a pinned \textbf{fMRIPrep 25.2.3}
Singularity container with fixed arguments, which makes preprocessing
reproducible. FreeSurfer surface reconstruction is disabled, since it dominates
runtime at this scale and the released representation is volumetric. The exact
command of every subject is stored in its provenance record. The released corpus consists of the subjects with
fMRIPrep output, after the run exclusions described in Data Records.

\paragraph{Step 2: Parcellation.} Each preprocessed run is parcellated under all
\nAtlases{} atlases from \nFamilies{} families (Table~\ref{tab:atlases}), a broad
selection of the atlases in common use. They include hard and soft parcellations,
anatomical and functional atlases, and cortical and subcortical coverage, with 10
to 1{,}024 regions, so that users can choose the atlas that suits their analysis.
Several families, such as Schaefer and Yan (26 of the 52 variants), come in
multiple resolutions and network assignments, which also allows studies of the
effect of resolution within a family. Hard parcellations use a labels masker and soft,
spatially overlapping decompositions use a maps masker, which avoids reducing
probabilistic atlases to winner-take-all assignments. The result is an
$(n_\text{timepoints} \times n_\text{ROIs})$ matrix for each run and atlas, with
no confound regression, scrubbing, filtering or standardization, so that the
timeseries stays frame-aligned with the fMRIPrep confounds file. This is the
\texttt{quasiraw} version. Runs are processed independently, as an array of
cluster jobs.

\paragraph{Step 3: Denoising.} The \texttt{preproc} version applies a default
denoising pipeline to the parcellated signal, in a single step that uses the
confounds file of the run. The pipeline (i) regresses out 26 fMRIPrep confounds
(the 24-parameter expanded motion model plus white matter and cerebrospinal
fluid signals); (ii) censors volumes with framewise displacement (FD) above
0.5~mm before filtering; (iii) applies linear detrending and a 0.01 to 0.1~Hz
Butterworth bandpass filter; and (iv) standardizes each ROI timeseries to unit
variance. The global signal is not regressed, and the censoring threshold is
permissive by design (see Motion and retention). The step depends only on
NumPy, pandas, nibabel and Nilearn and applies Nilearn's signal cleaning to the
timeseries. The sidecar of each
\texttt{preproc} file records what was applied. Every step is linear in time, so
denoising the parcellated signal is equivalent to denoising the voxelwise data
before parcellation. Users can therefore apply the same step to \texttt{quasiraw}
with other settings, for example with global signal regression or another FD
threshold, or use any other denoising tool. Runs with fewer than 50 volumes after
scrubbing are dropped from \texttt{preproc} as uninformative for sequence
modeling, so \nShortRuns{} runs are present in \texttt{quasiraw} only.

\paragraph{Step 4: Connectivity.} For both versions, functional connectivity is
computed per run and atlas as the Pearson correlation matrix of the ROI
timeseries (Nilearn's \texttt{ConnectivityMeasure}). In \texttt{preproc}, the
correlation uses the volumes retained after scrubbing.

\begin{table}[t]
  \caption{Atlas inventory, grouped by parcellation family. ``Hard'' atlases
  assign each voxel to exactly one parcel; ``soft'' atlases are probabilistic
  and spatially overlapping. ROI counts are those of the released sidecars
  (\texttt{ROILabels}).}
  \label{tab:atlases}
  \centering
  \small
  \begin{tabularx}{\linewidth}{llcX}
    \toprule
    Family & Variants & Type & Resolutions (ROIs) \\
    \midrule
    Schaefer \citep{schaefer2018} & 14 & Hard & 100 to 1000 (7n: 8 resolutions; 17n: 100, 200, 400, 1000); +Tian S2 (432), S3 (450) \\
    Yan            & 12 & Hard & 100 to 800 $\times$ \{yeo7, yeo17, kong17\} \\
    DiFuMo         & 5  & Soft & 64, 128, 256, 512, 1024 \\
    BASC           & 5  & Hard & 64, 122, 197, 325, 444 \\
    Smith          & 3  & Soft & 10, 20, 70 \\
    Tian \citep{tian2020} & 3 & Hard & 16 (S1), 32 (S2), 50 (S3); subcortical \\
    Harvard-Oxford & 2  & Hard & 48 (cortical), 21 (subcortical) \\
    Yeo            & 2  & Hard & 7, 17 networks \\
    AICHA          & 1  & Hard & 384 (homotopic) \\
    Glasser        & 1  & Hard & 360 (HCP MMP 1.0) \\
    AAL            & 1  & Hard & 166; anatomical \\
    Destrieux      & 1  & Hard & 150; anatomical \\
    MSDL           & 1  & Soft & 39 \\
    Pauli          & 1  & Hard & 16; subcortical \\
    \midrule
    \textbf{Total} & \textbf{\nAtlases} & & \textbf{\nFamilies{} families} \\
    \bottomrule
  \end{tabularx}
\end{table}

\section{Data Records}
\label{sec:records}

\dataset{} is hosted on Hugging Face at
\url{https://huggingface.co/datasets/cbarbano/open-fmind}.

\subsection{Contents}

The release comprises \relRuns{} functional runs from \effSubjects{} subjects in
\relDatasets{} OpenNeuro studies, in two versions of the same runs.
\texttt{quasiraw} contains all runs, and \texttt{preproc} the \prepRuns{} runs
(\prepSubjects{} subjects, \prepDatasets{} datasets) with at least 50 volumes
after scrubbing. Each run is parcellated under all \nAtlases{} atlases, which
yields \relSeriesQ{}~million 4D volumes in \texttt{quasiraw} (\relRuns{} runs
$\times$ \nAtlases{} atlases) and \relSeriesP{}~million in \texttt{preproc}
(\prepRuns{} $\times$ \nAtlases{}), each with its connectivity matrix. These
figures count run-atlas pairs. Every atlas view is a
deterministic function of the same BOLD run, and the two versions are functions
of the same fMRIPrep output, so the effective sample size for any subject- or
run-level analysis is \effSubjects{} subjects and \relRuns{} runs. The released datasets carry \relTasks{} unique
task labels (case-folded), and \relMultiSession{} of them are multi-session.
Table~\ref{tab:funnel} compares the release with the fMRI datasets available on
OpenNeuro. Of the \nKnownSubjects{} subjects retrieved 2,637 failed in fMRIPrep, which leaves \effSubjects{} subjects in
the release.

\begin{table}[t]
  \caption{Datasets, subjects and runs on OpenNeuro, in the data retrieved for
  this release, and in \dataset{}.}
  \label{tab:funnel}
  \centering
  \small
  \renewcommand{\arraystretch}{1.15}
  \begin{tabular}{lrrr}
    \toprule
     & Datasets & Subjects & Runs \\
    \midrule
    OpenNeuro (all fMRI datasets)               & \nDatasets   & \nSubjects      & \\
    Retrieved for this release                  & \nDownloaded & \nKnownSubjects & \nRawRuns \\
    \midrule
    \multicolumn{4}{l}{\textbf{Released} (excluding failures)} \\
    \quad \textbf{\dataset{}} (\texttt{quasiraw}) & \textbf{\relDatasets} & \textbf{\effSubjects} & \textbf{\relRuns} \\
    \quad \textbf{\dataset{}} (\texttt{preproc})  & \textbf{\prepDatasets} & \textbf{\prepSubjects} & \textbf{\prepRuns} \\
    \bottomrule
  \end{tabular}
\end{table}

\subsection{Organization and format}

Data follow BIDS derivative naming \citep{bids}, with one tree per kind and
version (\texttt{quasiraw} timeseries, connectivity and fMRIPrep confounds;
\texttt{preproc} timeseries and connectivity). Each run and atlas yields one
NIfTI file with a JSON sidecar:

\begin{quote}\ttfamily\small
<dataset>/sub-<ID>/[ses-<ID>/]func/sub-<ID>\_..\_atlas-<ATLAS>\_timeseries.nii.gz \\
<dataset>/sub-<ID>/[ses-<ID>/]func/sub-<ID>\_..\_atlas-<ATLAS>\_timeseries.json
\end{quote}

The timeseries has shape $n_\text{ROIs} \times 1 \times 1 \times
n_\text{timepoints}$ and the connectivity matrix $n_\text{ROIs} \times
n_\text{ROIs} \times 1 \times 1$, both in float32, so that standard neuroimaging
tools read them directly. The sidecar records the atlas name, the ordered ROI
labels, the repetition time, the confound model, the bandpass cutoffs, the FD
threshold, the number of volumes scrubbed, the mean and maximum FD, and the
processing chain with versions and command lines. It also records the
provenance of the run (source accession and dataset version, and fMRIPrep
container digest), so that any timeseries can be traced to the raw scan it
derives from. %
To keep the repository to a few thousand files, the files are packed into
\nShards{} uncompressed tar shards of about 3~GB (\releaseTB{}~TB in total), in
which every file keeps its BIDS path. Shards are split by atlas and by a coarse
\texttt{rest}/\texttt{task} label, so that resting-state data can be downloaded
alone. A parquet index lists every file with its shard and byte offset, so that
a single run can be read with an HTTP range request.

\subsection{Exclusions and known issues}

\nShortRuns{} runs are in \texttt{quasiraw} only, because fewer than 50 volumes
remained after scrubbing. Two datasets (ds000170, ds002675) are in
\texttt{quasiraw} only. Apart from these exclusions, no run was removed for
image quality or motion, and \texttt{quasiraw} is not scrubbed.

\subsection{Accompanying tables}

Tabular records accompany the imaging derivatives: the parquet \textbf{file
index} (one row per file); \texttt{LICENSES.tsv} (one row per source dataset,
with license, authors, DOI and requested acknowledgement); \texttt{atlases.tsv};
\texttt{task\_type\_mapping.tsv}, which gives the task label of every
(dataset, task) pair; and \texttt{qc.tsv}, a per-run table of motion. It has one
row for each of the \relRuns{} \texttt{quasiraw} runs (the \texttt{preproc} runs
are a subset), identified by dataset, subject, session, task, run and the other
BIDS entities, with the mean and maximum framewise displacement, the proportion
of volumes above the 0.5~mm threshold, and the number of volumes. These are
computed from the fMRIPrep confounds over all frames, so they describe the motion
of the run before scrubbing and are the same for both versions. The sidecar of
each run records the mean and maximum framewise displacement and the number of
volumes scrubbed.

\section{Technical Validation}
\label{sec:validation}

Technical Validation checks the quality of the imaging derivatives with
label-free analyses, and tests whether the corpus supports self-supervised
pretraining by evaluating a pretrained model on held-out datasets.

\subsection{Data quality}

The analyses read the connectivity matrices of the
released \texttt{preproc} version in place (from the parquet index and tar
shards), at Schaefer-400 unless stated, so that they test the files that users
download.

\subsubsection{Motion and retention}

For every released run we report mean and maximum FD, the proportion of volumes
above the 0.5~mm threshold, and the number of volumes. The
release threshold is permissive by design, since a pretraining corpus should let
downstream users choose their own trade-off between quality and quantity. Over
all released runs (FD computed over all frames from the fMRIPrep
confounds), the median of the per-run mean FD is 0.123~mm (interquartile range
0.088 to 0.181~mm; 95th percentile 0.408~mm). At the 0.5~mm threshold, $58.4\%$
of runs lose at least one volume to scrubbing, the median run loses $0.6\%$ of
its volumes, $5.6\%$ of runs lose more than $20\%$, and $1.0\%$ lose more than
half.

\subsubsection{Organization of runs in connectivity space}

\begin{figure}[tb]
  \centering
  \includegraphics[width=\linewidth]{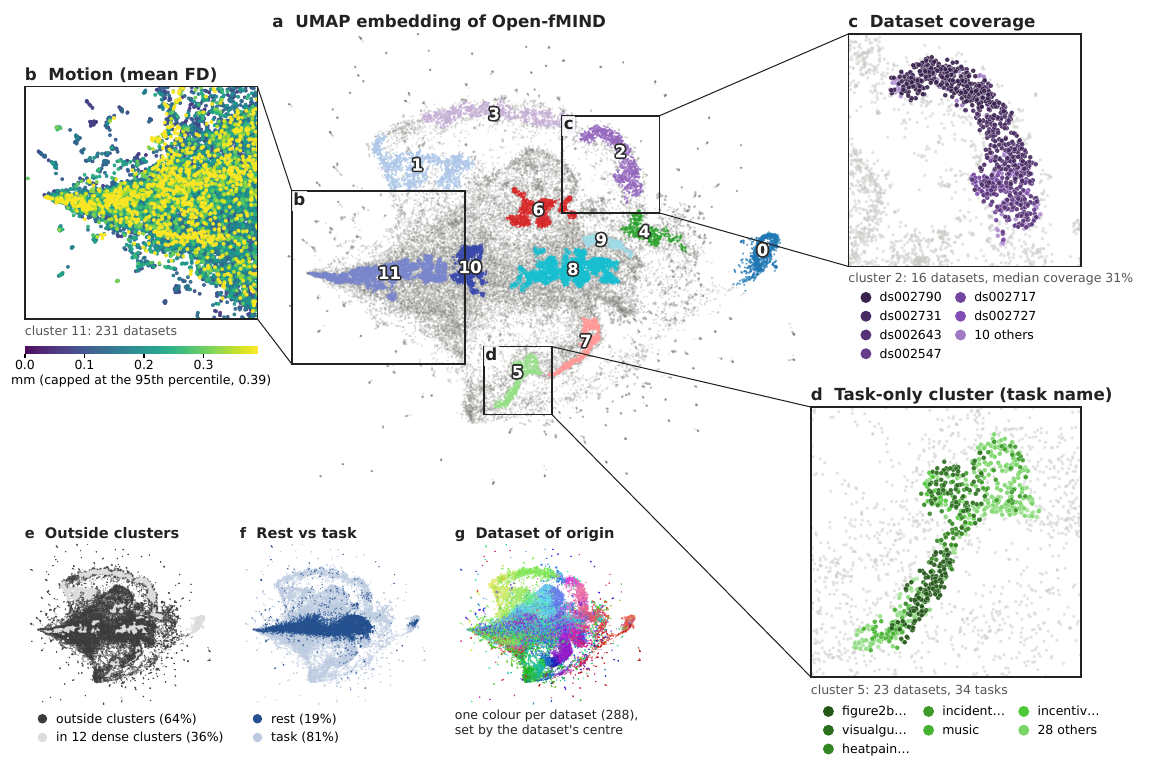}
  \caption{\textbf{Runs form regions in connectivity space that differ in
  motion, dataset of origin and task.} (a) UMAP (15 neighbors, minimum distance
  0.1) of the Schaefer-400 connectivity vectors of 44{,}241 preprocessed runs
  from 288 datasets; datasets with fewer than 5 subjects are left out (32
  datasets, 63 subjects, 778 runs). The 12 dense clusters found by HDBSCAN
  (minimum cluster size 500, minimum samples 50, leaf selection; 36\% of runs) are
  colored and numbered, and the other runs are grey. Boxes mark the clusters
  enlarged in (b-d), each of which maximizes one criterion. (b) Motion: the cluster
  with the highest median mean FD (cluster 11, 231 datasets; 0.165~mm against
  0.124~mm over all runs), colored by mean FD capped at the 95th percentile. (c)
  Dataset coverage: the cluster with the highest median dataset coverage
  (cluster 2, 16 datasets, 31\%), colored by dataset. (d) Task-only cluster: the
  cluster with the lowest share of rest runs (cluster 5, no rest runs, 23
  datasets, 34 task names), colored by task name. In (c) and (d), runs of other
  clusters inside the box are grey, and legends name the six most frequent
  groups. (e) The 64\% of runs outside any dense cluster. (f) Rest (19\%) and
  task (81\%) runs. (g) All runs colored by dataset. Each dataset
  has its own color, set by the median position of its runs in the plot (hue by
  x, lightness by y), so datasets with nearby centers have similar colors. Axes
  are arbitrary.}
  \label{fig:embedding}
\end{figure}

Fig.~\ref{fig:embedding} shows how runs are arranged in connectivity space and
what distinguishes its regions: motion, dataset of origin, or task. It uses only
BIDS metadata (dataset, task entity) and the FD of the QC table. The full
Schaefer-400 connectivity vector of every run (79{,}800 Fisher-$z$ edges) is
embedded in two dimensions with UMAP, and dense clusters are found with HDBSCAN.
Three clusters are enlarged, and each window maximizes one criterion computed
from the data: the amount of motion (highest median mean FD); the dataset
coverage (highest median \emph{dataset coverage}, defined for a dataset in a
cluster as the share of that dataset's runs that fall in the cluster and taken as
the median over the cluster's datasets); and the task content (lowest share of
rest runs). The preprocessed runs do not form one homogeneous cloud. Twelve dense clusters of 549 to 3{,}147 runs hold 36\% of
runs, and the other 64\% lie in the connected bulk between them
(Fig.~\ref{fig:embedding}a,e). Rest runs (19\%) concentrate in a band through the
embedding (Fig.~\ref{fig:embedding}f). The cluster with the highest median FD
(3{,}147 runs from 231 datasets) has a median mean FD of 0.165~mm against
0.124~mm over all runs (Fig.~\ref{fig:embedding}b), so higher motion is
concentrated in one region and shared across many datasets, which is consistent
with the residual motion signature in Fig.~\ref{fig:validation}m,n. Even in the
cluster with the highest dataset coverage, the median dataset has only 31\% of
its runs in the cluster, and across the twelve clusters the median coverage
ranges from 2.3\% to 30.7\% (Fig.~\ref{fig:embedding}c), so the runs of a dataset
are spread over several regions and no cluster corresponds to a single dataset.
One cluster contains no rest runs and is populated exclusively by task runs
(Fig.~\ref{fig:embedding}d). The clusters depend on the clustering settings, and task names are mostly
specific to a dataset, so the task-name coloring in Fig.~\ref{fig:embedding}d
separates datasets and does not show that the same task groups across datasets.

\subsubsection{Population coverage and validation experiments}

\begin{figure}[p]
  \centering
  \includegraphics[width=\linewidth,height=0.78\textheight,keepaspectratio]{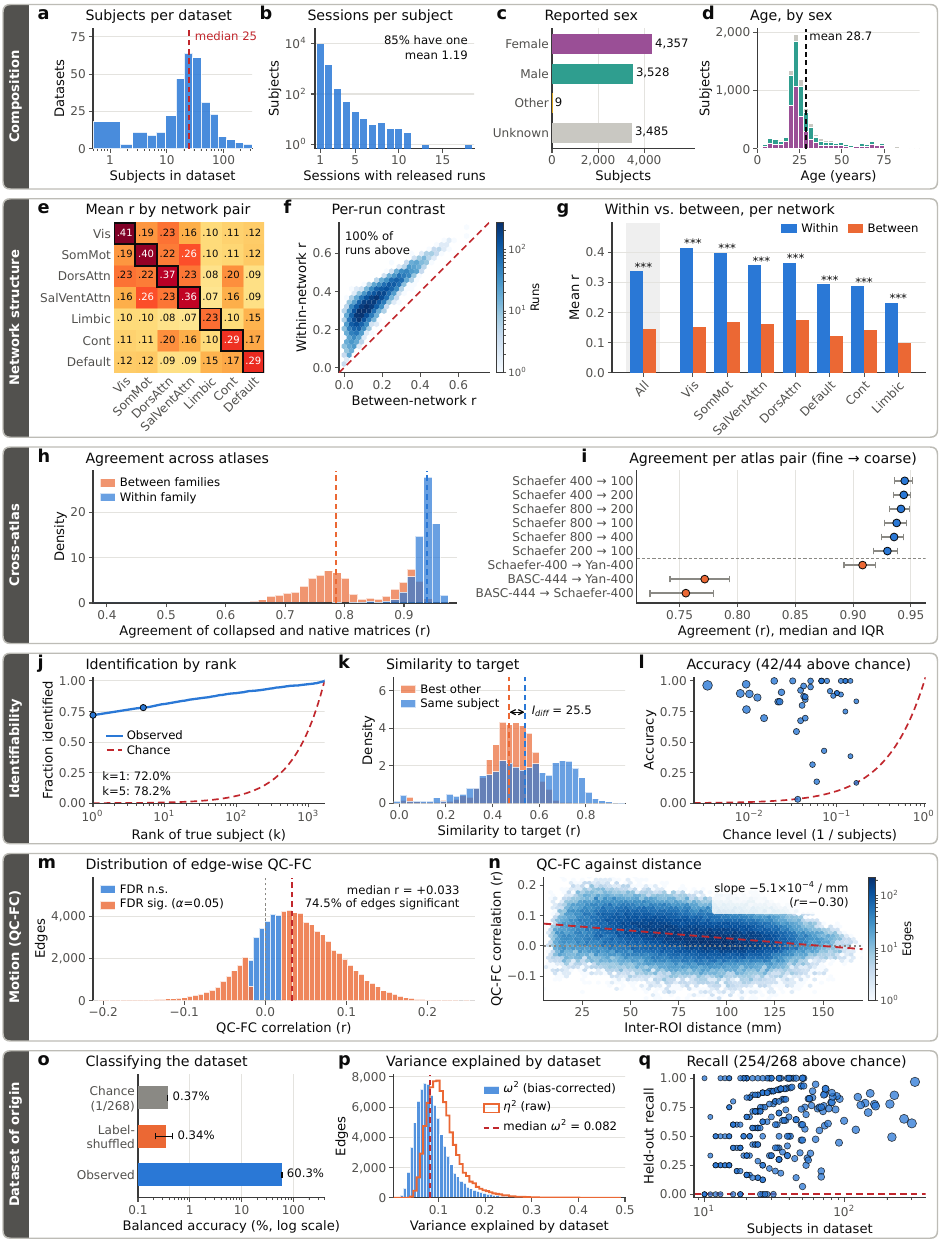}
  \captionsetup{font=small}
  \caption{\textbf{Composition, population coverage and validation experiments.}
  (a,b) Subjects per released dataset (dashed: median) and sessions per subject
  with released runs. (c,d) Reported sex and age (stacked by sex) of the
  released subjects. (e-g) Within- and between-network
  correlation in resting-state runs (Schaefer-400, Yeo 7 networks): mean per
  network pair (outlined diagonal: within-network), per run (dashed: equality),
  and per network (shaded: pooled; *** $p<0.001$, Holm-corrected). (h,i)
  Agreement of the finer atlas collapsed onto the coarser one with the native
  coarser matrix, per run (dashed: medians) and per atlas pair (median and
  interquartile range). (j-l) Identification across visits: fraction identified
  within the top $k$ (dashed: chance), similarity to the true and to the best
  other subject (dashed: means), and accuracy per dataset against chance
  (dashed: equality; marker area proportional to subjects). (m,n) QC-FC of
  79{,}800 edges by FDR significance (dashed red: median) and against inter-ROI
  distance (dashed red: least-squares fit). (o-q) Classification of the dataset
  of origin: balanced accuracy against chance and a permutation null, variance
  fraction per edge explained by dataset (dashed: median $\omega^2$), and recall
  per dataset against its size (dashed: chance). Panels e-q use the
  \texttt{preproc} version; sample sizes are in the text.}
  \label{fig:validation}
\end{figure}

Fig.~\ref{fig:validation} shows the composition and population coverage of the
corpus (panels a-d) and five experiments (panels e-q), each of which tests a
property that the data could fail to have. Edge correlations are
Fisher $z$-transformed before averaging.

\paragraph{Composition (a,b).} The released datasets contribute a
median of 25 subjects each (interquartile range 15 to 39; the largest has 317;
Fig.~\ref{fig:validation}a), and 85\% of subjects have one session with released
runs (Fig.~\ref{fig:validation}b).

\paragraph{Population coverage (c,d).} Coverage of age and sex is inherited from
the source studies. The \texttt{participants.tsv} files are not standardized, so
we resolve each column by name and content and discard ambiguous codes, such as
bare \texttt{0}/\texttt{1} for sex. \nDemogDatasets{} of the \relDatasets{} released
datasets publish a \texttt{participants.tsv}, covering \nDemogSubjects{} of the
\effSubjects{} released subjects. Percentages are of all \effSubjects{} released
subjects, so subjects without a file or a row count as unknown. Sex is
recoverable for \pctSexKnown\% of subjects (\pctFemale\% female, \pctMale\% male,
\pctOtherSex\% third-category, which is the German legal category \emph{Divers}
used in one dataset; Fig.~\ref{fig:validation}c) and age for \pctAgeKnown\%
(mean \meanAge, median \medianAge{} years), with a right skew toward young
adults (Fig.~\ref{fig:validation}d).

\paragraph{Network structure (e-g).} Denoising and parcellation should leave the
functional organization of the brain intact, so that within-network correlations
exceed between-network correlations in resting-state runs. A failure would
indicate a fault in confound regression, scrubbing or parcel assignment. We use
the 7-network Yeo labeling of Schaefer-400 and restrict the analysis to
resting-state runs. For each run, we average the edges inside networks and the
edges between networks, and compare the two means with a one-sided Wilcoxon
signed-rank test paired over runs, pooled and for each network, with Holm
correction. Over 8{,}704 runs from 5{,}219
subjects in 104 datasets, the mean within-network correlation ($r=0.34$) is more
than twice the mean between-network correlation ($r=0.15$;
Fig.~\ref{fig:validation}e,g). The difference holds for all seven networks (each
$p<0.001$ after Holm correction), for essentially every run ($100\%$ with within
above between; rank-biserial correlation $1.0$; Fig.~\ref{fig:validation}f), and
in all 104 datasets, so a few large studies do not carry the result.

\paragraph{Agreement across atlases (h,i).} Every run is parcellated under
\nAtlases{} atlases, so connectivity under different atlases should agree up to
the resolution of the parcellation. We use resting-state runs present under every atlas
compared. Within a family (Schaefer-100, 200, 400 and 800), each parcel of the
finer atlas is mapped to the coarser parcel that it overlaps most, and the finer
matrix is collapsed onto the coarser parcellation by averaging edges within each
mapped block. The collapsed matrix is then correlated with the natively computed
coarser matrix (Pearson $r$ over the upper triangle). The same procedure is
applied across families at matched resolution (Schaefer-400, BASC-444 and
Yan-400). Probabilistic atlases (DiFuMo, Smith, MSDL) are excluded, since mapping
them onto another parcellation needs a winner-take-all assignment that removes
their overlap. Across 500 runs from 464 subjects and 9 atlas pairs (6
within-family, 3 between-family), within-family agreement (median $r=0.939$)
exceeds between-family agreement (median $r=0.786$; one-sided Mann-Whitney $U$
test, $p<10^{-300}$; Fig.~\ref{fig:validation}h). A random within-family
comparison agrees more than a random between-family one $96.2\%$ of the time, and
all 6 within-family pairs (median agreement $0.930$ to $0.945$) lie above the
highest between-family pair (Schaefer-400 and Yan-400, $0.908$;
Fig.~\ref{fig:validation}i). Connectivity structure is
therefore preserved across atlases up to the coarseness of the parcellation,
which supports treating the atlas axis as a view or augmentation.

\paragraph{Subject identifiability (j-l).} The corpus should keep individual
variability after denoising, since chance-level identification would suggest that
denoising flattens differences between subjects. We test whether the connectivity
profile of a subject can be identified across separate visits \citep{finn2015}.
Each subject with at least two visits (one visit per session, since two runs of
one session share head position and shim)
places their first visit in a database, and their other visits are queries. A
query is correctly identified when its true subject ranks first among all
database entries by Pearson correlation of the Fisher-$z$ connectivity vectors.
We report a corpus-wide scope, where the database spans every subject, and a
within-dataset scope, where it is limited to the subjects of the same OpenNeuro
accession, which rules out matching on scanner characteristics. We also report
differential identifiability $I_\text{diff}$, the mean within-subject similarity
minus the mean between-subject similarity, in correlation points $\times 100$
\citep{finn2015}. Across 1{,}649 subjects with at least two visits (2{,}192 query
visits) in 57 datasets, corpus-wide accuracy is $72.0\%$ against a chance level
of $0.06\%$ ($1{,}187\times$ enrichment), and top-5 accuracy is $78.2\%$
(Fig.~\ref{fig:validation}j). The within-dataset scope covers the 44 datasets
with enough multi-visit subjects (2{,}104 queries) and gives $85.1\%$ accuracy
against a mean chance of $2.7\%$, with 42 of the 44 datasets above their own
chance level (Fig.~\ref{fig:validation}l). The similarity to the true subject
exceeds that to the best other subject by $I_\text{diff}=25.5$ correlation points
(Fig.~\ref{fig:validation}k). Two visits of one subject often share a task
($94.0\%$ of queries), so part of the signal could reflect the task: accuracy is
$73.6\%$ when the task is shared and $47.0\%$ when it is not, both far above
chance.

\paragraph{Residual motion (m,n).} Connectivity should reflect neural signal with
little residual head motion after confound regression and scrubbing. For each
edge, we correlate connectivity strength with the mean FD of the subject (QC-FC),
using one run per subject and the mean FD over all frames from the fMRIPrep
confounds. A well-denoised release shows QC-FC correlations near zero and no
dependence on the distance between the two ROIs, since distance dependence is the
signature of motion artifact that regression failed to remove
\citep{power2012,parkes2018}. We assess significance with
Benjamini-Hochberg control of the false discovery rate (FDR) at $\alpha=0.05$
and fit QC-FC linearly against the distance between parcel centroids. Across
11{,}319 subjects (one run per subject) from 319 datasets,
the 79{,}800 edges show a QC-FC distribution centered slightly above zero (median
$r=+0.033$, median $|r|=0.042$; Fig.~\ref{fig:validation}m), and $74.5\%$ of
edges are FDR-significant. At this sample size even small effects are
significant, so the size of the effects is a more informative summary than the
fraction of significant edges \citep{parkes2018}. QC-FC decreases with inter-ROI
distance (slope $-5.07\times10^{-4}$ per~mm, $r=-0.296$, $p<10^{-300}$;
Fig.~\ref{fig:validation}n), which is the classical signature of motion
artifact. The released matrices therefore carry a modest residual motion
signature that is statistically robust. It is small in absolute terms: the fitted
slope predicts a change of about 0.08 correlation points over the roughly 165~mm
distance range of the atlas.

\paragraph{Dataset of origin (o-q).} This experiment asks how much of the
variance in connectivity is attributable to the source dataset (scanner, sequence
and site), so that users can decide whether harmonization (for example ComBat) is
needed. We use one run per subject, so that a dataset with many runs per subject
cannot inflate its own separability, and drop datasets with fewer than 10
subjects. A multinomial logistic regression predicts the dataset of origin from
the vectorized upper triangle, on a stratified 70/30 split made by subject;
features are standardized and reduced by PCA (200 components) fitted on the
training split only. Balanced accuracy is compared with the theoretical chance
level ($1/n_\text{classes}$) and with a label-permutation null (20 permutations).
As a view that does not depend on a classifier, a one-way ANOVA per edge with the
dataset of origin as factor gives the between-dataset variance fraction as the
bias-corrected $\omega^2$. Across 11{,}117 subjects from
268 datasets (52 smaller datasets dropped), balanced accuracy is $60.3\%$ against
a theoretical chance of $0.37\%$ ($162\times$ enrichment) and a permutation null
of $0.34\%\pm0.12\%$ (Fig.~\ref{fig:validation}o). 254 of the 268 datasets are individually classified above
chance, so a few large studies do not carry the result, and held-out recall
increases gradually with dataset size (Fig.~\ref{fig:validation}q). The
edge-level decomposition agrees with the classifier: the median $\omega^2$ is
$0.082$, so the dataset of origin
accounts for roughly 8\% of the between-subject variance of a typical edge, and
$89.6\%$ of edges exceed $\omega^2=0.05$ (Fig.~\ref{fig:validation}p). A site
effect of this size is expected for a corpus built from independently designed
studies.

\subsection{Pretraining on \dataset{}}

We test whether \dataset{} supports the pretraining of a published fMRI
foundation model, Brain-Semantoks \citep{semantoks}. Three frozen encoders are
compared on the same downstream tasks: the model pretrained on \dataset{}, the
checkpoint released by its authors, which is pretrained on UK Biobank, and the
same architecture with random weights.

\paragraph{Pretraining.} Brain-Semantoks groups the Schaefer-400 regions into the
7 Yeo networks and encodes one token per network with an 8-layer transformer of
width 768. It is trained by self-distillation: a student and a teacher network see
different 100-volume crops of the same run, and on half of the steps most of the
student's network tokens are masked. We use the implementation and hyperparameters
of the authors (100 epochs, batch size 512, one GPU) and evaluate the epoch-50
checkpoint, a choice fixed before any evaluation. The input is the Schaefer-400
\texttt{preproc} timeseries, resampled to a repetition time of 2~s, cut into
windows of 160 volumes and z-scored per region within each window. Runs shorter
than 320~s are excluded, which leaves 51{,}244 windows from 29{,}817 runs of
9{,}662 subjects in 284 datasets. We do not use the subcortical and cerebellar
streams of the original model. The released UK Biobank checkpoint uses both, and
since our evaluation data have no cerebellar parcellation, its cerebellar channels
are set to zero.

\paragraph{Evaluation.} For each subject, the frozen teacher encodes windows of
the timeseries, and the class token concatenated with the mean network token
(1{,}536 features) is averaged over windows. All encoders use the same subjects,
folds and classifiers. On HCP-YA \citep{hcp} (955 subjects, 422 families), we
predict the general cognition factor with kernel ridge regression, scored by
Pearson $r$ with the nested cross-validation of Ooi et al.~\citep{ooi2022} (20 repetitions
of 10 folds, families kept within a fold), and sex with logistic regression,
scored by the area under the ROC curve (AUROC) over 20 repetitions of family-aware
5-fold cross-validation. On ABIDE \citep{abide} (up to 1{,}035 subjects) and
ADHD-200 \citep{adhd200} (up to 298 subjects), we predict diagnosis, sex, and age
in 5 classes of equal size with class-balanced logistic regression, scored by
balanced accuracy over 10 repetitions of stratified 5-fold cross-validation.

\begin{figure}[tb]
  \centering
  \includegraphics[width=\linewidth]{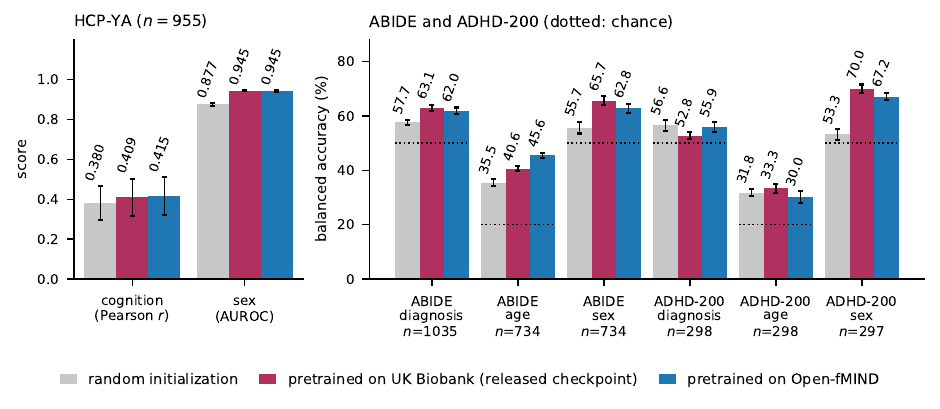}
  \caption{\textbf{A model pretrained on \dataset{} reaches accuracy close to
  that of a model pretrained on UK Biobank.} Frozen Brain-Semantoks encoders with
  a linear or kernel readout on HCP-YA, ABIDE and ADHD-200. Bars are means
  and error bars standard deviations over cross-validation folds (200 for HCP
  cognition) or repetitions (20 for HCP sex, 10 for ABIDE and ADHD-200). Accuracies
  are balanced accuracies, age is predicted as one of 5 classes of equal size, and
  dotted lines mark chance (50\% for two classes, 20\% for five).}
  \label{fig:pretraining}
\end{figure}

\paragraph{Results.} On HCP-YA, the model pretrained on \dataset{} matches the one
pretrained on UK Biobank for sex (AUROC $0.945$ for both) and cognition ($r=0.415$
against $0.409$, with a standard deviation of about $0.095$ across folds), and
both exceed random initialization ($0.877$ and $0.380$;
Fig.~\ref{fig:pretraining}). On ABIDE and ADHD-200, the \dataset{} model is about
1 point below the UK Biobank model for ABIDE diagnosis ($62.0\%$ against
$63.1\%$), about 3 points below for sex ($62.8\%$ against $65.7\%$ on ABIDE,
$67.2\%$ against $70.0\%$ on ADHD-200), and 5 points above for ABIDE age
($45.6\%$ against $40.6\%$). For sex and for ABIDE diagnosis and age, both
pretrained models are above random initialization. For ADHD-200 diagnosis and
age, no model is clearly better than random initialization. A model pretrained on
\dataset{} alone therefore reaches accuracy close to that of a model pretrained
on a gated resource. Each encoder comes from a single pretraining run, and the
error bars reflect cross-validation resampling only, so small differences should
be interpreted with care.

\section{Usage Notes}
\label{sec:usage}

\subsection{Intended use}

\dataset{} is built as a pretraining resource. The ROI timeseries can be used
directly as input sequences for sequence models, and the corpus supports
self-supervised objectives (masked reconstruction, contrastive, predictive) that
need no labels. It has been used for pretraining in other works, for
brain-behaviour prediction with prior-data fitted networks
\citep{barbano2026brainpfn} and for fMRI encoders trained with a spectral
connectivity target \citep{marraffini2026flattening}. Users pursuing supervised tasks need to curate targets from the
source studies' own metadata, using the provenance records to trace each run to
its accession. We advise against assembling phenotypes naively across studies,
since identical column names in different \texttt{participants.tsv} files do not
imply identical instruments, units, or coding.

The \nAtlases{} atlases give each run \nAtlases{} representations. Users can
quantify the sensitivity of a result to parcellation choice, or treat the
parcellation as a view, for instance with contrastive objectives that draw
positive pairs from two atlas views of the same run. Two practical points apply.
Variants of one family are more similar to one another than to other atlases, and
Schaefer and Yan make up half of all variants (Table~\ref{tab:atlases}), so
sampling a family first and then a variant spreads the views more evenly than
sampling atlases uniformly. Atlas views are also deterministic transforms of one
run and carry no independent information about the subject, so they add
representational diversity and leave the sample size unchanged.

\subsection{Caveats}

\textbf{Site and scanner effects.} The corpus aggregates
independently designed studies with different scanners, sequences, and
acquisition parameters. Site effects are present and users should account for
them. The provenance records identify the source dataset of every run, and
Fig.~\ref{fig:validation}o-q quantify the effect.
\textbf{Heterogeneity.} Runs span resting-state, cognitive, naturalistic, and
clinical paradigms with no common protocol. We consider this an asset for general
representation learning, and it makes comparison with single-protocol resources
harder.
\textbf{Parcellated release.} Only ROI timeseries are released, so analyses
that need fine spatial detail are out of scope, although the pipeline can
regenerate other atlases from the fMRIPrep outputs.
\textbf{Population coverage.} Age and sex coverage is uneven and inherited
from the source studies (Fig.~\ref{fig:validation}c,d), so users should not assume that the
corpus represents any population.
\textbf{Motion.} The FD threshold is permissive, and Fig.~\ref{fig:validation}m,n
show a modest residual motion signature. Users with stricter requirements should filter
on the per-run motion metrics in \texttt{qc.tsv}.

\subsection{Licensing and ethics}

All source data are shared openly on OpenNeuro~\citep{\releasedCites}. The derivatives of each dataset
keep the license of their source, together with its provenance. The released
datasets carry CC0-1.0 (251), PDDL-1.0 (34), CC-BY-SA-4.0 (3), CC-BY-4.0 (2),
CC-BY-NC-4.0 (1) and one other license, and 30 state no license in their
OpenNeuro metadata. \texttt{LICENSES.tsv} lists the license, authors, DOI and
requested acknowledgement of every dataset, and Appendix~\ref{app:datasets}
lists the datasets with their names, licenses and citations. All of these licenses permit the
redistribution of derivatives. The CC-BY-SA-4.0 datasets require derivatives to
be shared under the same license, and the CC-BY-NC-4.0 dataset restricts its
derivatives to non-commercial use. The 30 datasets without a stated license are
treated as CC0, following the terms under which datasets are uploaded to
OpenNeuro. We will remove any dataset at the request of its authors.

The underlying data were collected under the original studies' ethical approvals
and shared as de-identified, defaced BIDS data. We performed no new data
collection, and no re-identification was attempted at any stage.

\section{Code Availability}
\label{sec:code}

The complete pipeline (dataset discovery, retrieval, fMRIPrep submission,
denoising and parcellation, QC collection, and reporting) is released as an
open-source repository, together with the job script of each fMRIPrep run and the
denoising script. The repository is available at
\url{https://github.com/carloalbertobarbano/open-fmind}.
The environment is specified by a lockfile, and preprocessing runs from a pinned
fMRIPrep 25.2.3 \citep{fmriprep} Singularity container. Atlases are fetched
through Nilearn \citep{nilearn} and from their original distributions, and are
resolved by a single registry in \texttt{parcellate\_bold.py}.

\section*{Acknowledgements}
This work was supported by the French National Research Agency (ANR) through
the MICBrainPres project (grant ANR-22-CE19-0013).
This work was granted access to the HPC resources of IDRIS under the allocation
2026-AD011017736 and 2026-AD010617904 made by GENCI.

\section*{Author Contributions}
C.A.B. conceived the dataset, built the pipeline, ran the preprocessing and
parcellation, performed the technical validation and wrote the manuscript.
G.M. contributed to building the pipeline and ran the deep learning validation.
D.W. supervised the project, acquired funding and reviewed the manuscript.
All authors read and approved the final manuscript.

\section*{Competing Interests}
The authors declare no competing interests.

\bibliographystyle{unsrtnat}
\bibliography{refs,openneuro_datasets}

\newpage
\appendix
\section{Released datasets}
\label{app:datasets}

Table~\ref{tab:released-datasets} lists the \relDatasets{} OpenNeuro datasets
with released runs.

\input{released_datasets}

\end{document}

%% file: released_cites.tex
\newcommand{\releasedCites}{%
ds000001,%
ds000002,%
ds000003,%
ds000005,%
ds000006,%
ds000007,%
ds000008,%
ds000009,%
ds000011,%
ds000017,%
ds000030,%
ds000101,%
ds000102,%
ds000107,%
ds000108,%
ds000109,%
ds000110,%
ds000113,%
ds000114,%
ds000115,%
ds000116,%
ds000119,%
ds000120,%
ds000121,%
ds000133,%
ds000140,%
ds000144,%
ds000148,%
ds000149,%
ds000157,%
ds000168,%
ds000170,%
ds000171,%
ds000172,%
ds000200,%
ds000203,%
ds000205,%
ds000208,%
ds000210,%
ds000212,%
ds000213,%
ds000214,%
ds000218,%
ds000219,%
ds000220,%
ds000221,%
ds000223,%
ds000228,%
ds000229,%
ds000231,%
ds000233,%
ds000243,%
ds000245,%
ds000249,%
ds000253,%
ds000254,%
ds000256,%
ds000258,%
ds001021,%
ds001131,%
ds001132,%
ds001145,%
ds001226,%
ds001229,%
ds001232,%
ds001235,%
ds001241,%
ds001242,%
ds001247,%
ds001297,%
ds001302,%
ds001338,%
ds001344,%
ds001345,%
ds001353,%
ds001357,%
ds001379,%
ds001399,%
ds001408,%
ds001419,%
ds001430,%
ds001439,%
ds001454,%
ds001491,%
ds001521,%
ds001534,%
ds001545,%
ds001553,%
ds001554,%
ds001566,%
ds001576,%
ds001607,%
ds001608,%
ds001612,%
ds001621,%
ds001634,%
ds001635,%
ds001715,%
ds001722,%
ds001728,%
ds001740,%
ds001745,%
ds001747,%
ds001748,%
ds001762,%
ds001775,%
ds001796,%
ds001838,%
ds001840,%
ds001847,%
ds001907,%
ds001912,%
ds001921,%
ds001923,%
ds001927,%
ds001928,%
ds002000,%
ds002011,%
ds002013,%
ds002014,%
ds002041,%
ds002080,%
ds002149,%
ds002232,%
ds002250,%
ds002270,%
ds002322,%
ds002345,%
ds002363,%
ds002366,%
ds002522,%
ds002547,%
ds002603,%
ds002608,%
ds002609,%
ds002643,%
ds002675,%
ds002711,%
ds002715,%
ds002717,%
ds002727,%
ds002731,%
ds002732,%
ds002734,%
ds002737,%
ds002748,%
ds002750,%
ds002785,%
ds002790,%
ds002793,%
ds002797,%
ds002835,%
ds002837,%
ds002841,%
ds002842,%
ds002843,%
ds002848,%
ds002872,%
ds002896,%
ds002938,%
ds002940,%
ds003007,%
ds003011,%
ds003012,%
ds003037,%
ds003076,%
ds003083,%
ds003085,%
ds003089,%
ds003096,%
ds003103,%
ds003114,%
ds003126,%
ds003146,%
ds003171,%
ds003342,%
ds003346,%
ds003354,%
ds003381,%
ds003382,%
ds003425,%
ds003427,%
ds003436,%
ds003454,%
ds003459,%
ds003466,%
ds003469,%
ds003487,%
ds003495,%
ds003500,%
ds003507,%
ds003542,%
ds003545,%
ds003548,%
ds003550,%
ds003553,%
ds003554,%
ds003569,%
ds003574,%
ds003592,%
ds003688,%
ds003709,%
ds003763,%
ds003770,%
ds003772,%
ds003789,%
ds003831,%
ds003835,%
ds003849,%
ds003858,%
ds003871,%
ds003892,%
ds003965,%
ds003972,%
ds003990,%
ds003999,%
ds004021,%
ds004101,%
ds004128,%
ds004134,%
ds004141,%
ds004142,%
ds004144,%
ds004182,%
ds004196,%
ds004213,%
ds004219,%
ds004302,%
ds004392,%
ds004400,%
ds004440,%
ds004450,%
ds004467,%
ds004469,%
ds004498,%
ds004544,%
ds004557,%
ds004562,%
ds004564,%
ds004594,%
ds004604,%
ds004627,%
ds004648,%
ds004656,%
ds004662,%
ds004697,%
ds004718,%
ds004720,%
ds004725,%
ds004737,%
ds004765,%
ds004775,%
ds004783,%
ds004798,%
ds004835,%
ds004848,%
ds004873,%
ds004956,%
ds005016,%
ds005027,%
ds005040,%
ds005069,%
ds005072,%
ds005075,%
ds005128,%
ds005134,%
ds005166,%
ds005230,%
ds005239,%
ds005339,%
ds005355,%
ds005365,%
ds005418,%
ds005454,%
ds005455,%
ds005479,%
ds005492,%
ds005504,%
ds005518,%
ds005525,%
ds005530,%
ds005533,%
ds005573,%
ds005588,%
ds005596,%
ds005598,%
ds005603,%
ds005623,%
ds005625,%
ds005699,%
ds005754,%
ds005783,%
ds005795,%
ds005850,%
ds005882,%
ds005883,%
ds005884,%
ds005892,%
ds005903,%
ds006092,%
ds006156,%
ds006193,%
ds006211,%
ds006559,%
ds006568,%
ds006572,%
ds006592,%
ds006594,%
ds006729,%
ds006893,%
ds006944,%
ds006945,%
ds006952,%
ds007115,%
ds007126,%
ds007135,%
ds007185,%
ds007313,%
ds007345,%
ds007386,%
ds007441,%
ds007534}

%% file: released_datasets.tex
{\footnotesize\captionsetup{font=normalsize}
\renewcommand{\arraystretch}{1.1}
\begin{longtable}{@{}l p{0.5\linewidth} l r r@{}}
\caption{OpenNeuro datasets in \dataset{} v2.0 (322 datasets), with the title, license and citation of the OpenNeuro snapshot that was downloaded (the latest one for the 17 datasets with no recorded snapshot; retrieved 2026-11-10). Citations give the authors, title, snapshot version and, when registered, the DOI. Subjects and Runs are the released subjects and runs of each dataset (\texttt{quasiraw}, which contains all released runs). Each accession links to its page at \texttt{https://openneuro.org/datasets/<accession>}. Licenses: CC0-1.0 (251), PDDL-1.0 (34), not stated (30), CC-BY-SA-4.0 (3), CC-BY-4.0 (2), CC-BY-NC-4.0 (1), other (1).}
\label{tab:released-datasets} \\
\toprule
Accession & Name & License & Subjects & Runs \\
\midrule
\endfirsthead
\toprule
Accession & Name & License & Subjects & Runs \\
\midrule
\endhead
\bottomrule
\endfoot
\href{https://openneuro.org/datasets/ds000001}{ds000001} & Balloon Analog Risk-taking Task \citep{ds000001} & CC0-1.0 & 16 & 48 \\
\href{https://openneuro.org/datasets/ds000002}{ds000002} & Classification learning \citep{ds000002} & PDDL-1.0 & 17 & 102 \\
\href{https://openneuro.org/datasets/ds000003}{ds000003} & Rhyme judgment \citep{ds000003} & CC0-1.0 & 13 & 13 \\
\href{https://openneuro.org/datasets/ds000005}{ds000005} & Mixed-gambles task \citep{ds000005} & PDDL-1.0 & 16 & 48 \\
\href{https://openneuro.org/datasets/ds000006}{ds000006} & ds000006 \citep{ds000006} & PDDL-1.0 & 14 & 162 \\
\href{https://openneuro.org/datasets/ds000007}{ds000007} & Stop-signal task with spoken \& manual responses \citep{ds000007} & PDDL-1.0 & 20 & 118 \\
\href{https://openneuro.org/datasets/ds000008}{ds000008} & Stop-signal task with unconditional and conditional stopping \citep{ds000008} & PDDL-1.0 & 14 & 78 \\
\href{https://openneuro.org/datasets/ds000009}{ds000009} & The generality of self-control \citep{ds000009} & PDDL-1.0 & 24 & 144 \\
\href{https://openneuro.org/datasets/ds000011}{ds000011} & Classification learning and tone-counting \citep{ds000011} & CC0-1.0 & 14 & 84 \\
\href{https://openneuro.org/datasets/ds000017}{ds000017} & ds000017 \citep{ds000017} & PDDL-1.0 & 8 & 74 \\
\href{https://openneuro.org/datasets/ds000030}{ds000030} & UCLA Consortium for Neuropsychiatric Phenomics LA5c Study \citep{ds000030} & CC0-1.0 & 265 & 1{,}969 \\
\href{https://openneuro.org/datasets/ds000101}{ds000101} & Simon task \citep{ds000101} & PDDL-1.0 & 21 & 42 \\
\href{https://openneuro.org/datasets/ds000102}{ds000102} & Flanker task (event-related) \citep{ds000102} & PDDL-1.0 & 26 & 52 \\
\href{https://openneuro.org/datasets/ds000107}{ds000107} & Word and object processing \citep{ds000107} & PDDL-1.0 & 49 & 96 \\
\href{https://openneuro.org/datasets/ds000108}{ds000108} & Prefrontal-Subcortical Pathways Mediating Successful Emotion Regulation \citep{ds000108} & PDDL-1.0 & 34 & 200 \\
\href{https://openneuro.org/datasets/ds000109}{ds000109} & Social cognitive tasks (animate movement task, moral judgment task, false belief task) \citep{ds000109} & CC0-1.0 & 36 & 69 \\
\href{https://openneuro.org/datasets/ds000110}{ds000110} & Incidental encoding task (Posner Cueing Paradigm) \citep{ds000110} & PDDL-1.0 & 18 & 179 \\
\href{https://openneuro.org/datasets/ds000113}{ds000113} & Forrest Gump \citep{ds000113} & not stated & 11 & 186 \\
\href{https://openneuro.org/datasets/ds000114}{ds000114} & A test-retest fMRI dataset for motor, language and spatial attention functions. \citep{ds000114} & CC0-1.0 & 10 & 100 \\
\href{https://openneuro.org/datasets/ds000115}{ds000115} & Working memory in healthy and schizophrenic individuals \citep{ds000115} & not stated & 99 & 297 \\
\href{https://openneuro.org/datasets/ds000116}{ds000116} & Auditory and Visual Oddball EEG-fMRI \citep{ds000116} & not stated & 17 & 101 \\
\href{https://openneuro.org/datasets/ds000119}{ds000119} & Maturational Changes in Anterior Cingulate and Frontoparietal Recruitment Support the Development of Error Processing and Inhibitory Control (Antistate) \citep{ds000119} & PDDL-1.0 & 72 & 279 \\
\href{https://openneuro.org/datasets/ds000120}{ds000120} & Developmental changes in brain function underlying the influence of reward processing on inhibitory control (Slot Reward) \citep{ds000120} & PDDL-1.0 & 26 & 70 \\
\href{https://openneuro.org/datasets/ds000121}{ds000121} & Immaturities in Reward Processing and Its Influence on Inhibitory Control in Adolescence (Ring Reward) \citep{ds000121} & PDDL-1.0 & 28 & 106 \\
\href{https://openneuro.org/datasets/ds000133}{ds000133} & Modafinil alters intrinsic functional connectivity of the right posterior insula: a pharmacological resting state fMRI study \citep{ds000133} & PDDL-1.0 & 26 & 156 \\
\href{https://openneuro.org/datasets/ds000140}{ds000140} & Distinct brain systems mediate the effects of nociceptive input and self-regulation on pain \citep{ds000140} & PDDL-1.0 & 33 & 297 \\
\href{https://openneuro.org/datasets/ds000144}{ds000144} & Preschool Anxiety Disorders \citep{ds000144} & CC0-1.0 & 45 & 66 \\
\href{https://openneuro.org/datasets/ds000148}{ds000148} & Who can afford self-control? The neural efficiency mechanism explains effective self-regulation of behavior \citep{ds000148} & PDDL-1.0 & 49 & 147 \\
\href{https://openneuro.org/datasets/ds000149}{ds000149} & Scene Perception \citep{ds000149} & CC0-1.0 & 15 & 44 \\
\href{https://openneuro.org/datasets/ds000157}{ds000157} & Block design food and nonfood picture viewing task \citep{ds000157} & CC0-1.0 & 30 & 30 \\
\href{https://openneuro.org/datasets/ds000168}{ds000168} & Offline Processing in Associative Learning \citep{ds000168} & PDDL-1.0 & 35 & 35 \\
\href{https://openneuro.org/datasets/ds000170}{ds000170} & Learning and memory: motor skill consolidation and intermanual transfer \citep{ds000170} & PDDL-1.0 & 15 & 179 \\
\href{https://openneuro.org/datasets/ds000171}{ds000171} & Neural Processing of Emotional Musical and Nonmusical Stimuli in Depression \citep{ds000171} & PDDL-1.0 & 39 & 195 \\
\href{https://openneuro.org/datasets/ds000172}{ds000172} & Physiological Contribution in Spontaneous Oscillations: An Approximate Quality - Assurance Index for Resting-State fMRI Signals \citep{ds000172} & PDDL-1.0 & 13 & 52 \\
\href{https://openneuro.org/datasets/ds000200}{ds000200} & Pre-adolescents Exposure to Manganese \citep{ds000200} & not stated & 13 & 13 \\
\href{https://openneuro.org/datasets/ds000203}{ds000203} & Visual imagery and false memory for pictures \citep{ds000203} & PDDL-1.0 & 26 & 52 \\
\href{https://openneuro.org/datasets/ds000205}{ds000205} & Affective Videos \citep{ds000205} & PDDL-1.0 & 11 & 33 \\
\href{https://openneuro.org/datasets/ds000208}{ds000208} & Brain connectivity predicts placebo response across chronic pain clinical trials \citep{ds000208} & CC0-1.0 & 76 & 76 \\
\href{https://openneuro.org/datasets/ds000210}{ds000210} & Multi-echo fMRI replication sample of autobiographical memory, prospection and theory of mind reasoning tasks \citep{ds000210} & PDDL-1.0 & 31 & 155 \\
\href{https://openneuro.org/datasets/ds000212}{ds000212} & Moral judgments of intentional and accidental moral violations across Harm and Purity domains \citep{ds000212} & CC0-1.0 & 39 & 372 \\
\href{https://openneuro.org/datasets/ds000213}{ds000213} & Neural mechanism underlying appearance social comparison \citep{ds000213} & PDDL-1.0 & 26 & 26 \\
\href{https://openneuro.org/datasets/ds000214}{ds000214} & EUPD Cyberball \citep{ds000214} & not stated & 36 & 36 \\
\href{https://openneuro.org/datasets/ds000218}{ds000218} & Flavour Pleasantness (Oral Nutritional Supplements) \citep{ds000218} & CC0-1.0 & 19 & 57 \\
\href{https://openneuro.org/datasets/ds000219}{ds000219} & Flavour Pleasantness (Regular Products) \citep{ds000219} & CC0-1.0 & 21 & 78 \\
\href{https://openneuro.org/datasets/ds000220}{ds000220} & Cost Analysis TBI \citep{ds000220} & CC0-1.0 & 26 & 66 \\
\href{https://openneuro.org/datasets/ds000221}{ds000221} & MPI-Leipzig\_Mind-Brain-Body \citep{ds000221} & CC0-1.0 & 317 & 1{,}014 \\
\href{https://openneuro.org/datasets/ds000223}{ds000223} & Magnitude Effect \citep{ds000223} & PDDL-1.0 & 19 & 78 \\
\href{https://openneuro.org/datasets/ds000228}{ds000228} & MRI data of 3-12 year old children and adults during viewing of a short animated film \citep{ds000228} & CC0-1.0 & 155 & 155 \\
\href{https://openneuro.org/datasets/ds000229}{ds000229} & Integration of sweet taste and metabolism determines carbohydrate reward - study 1 \citep{ds000229} & CC0-1.0 & 15 & 44 \\
\href{https://openneuro.org/datasets/ds000231}{ds000231} & Integration of sweet taste and metabolism determines carbohydrate reward-study 3 \citep{ds000231} & CC0-1.0 & 9 & 27 \\
\href{https://openneuro.org/datasets/ds000233}{ds000233} & Neural responses to naturalistic clips of behaving animals in two different task contexts \citep{ds000233} & CC0-1.0 & 12 & 120 \\
\href{https://openneuro.org/datasets/ds000243}{ds000243} & Washington University 120 \citep{ds000243} & CC0-1.0 & 120 & 203 \\
\href{https://openneuro.org/datasets/ds000245}{ds000245} & ds000245\_R1.0.0 \citep{ds000245} & CC0-1.0 & 45 & 45 \\
\href{https://openneuro.org/datasets/ds000249}{ds000249} & Value generalization in human avoidance learning \citep{ds000249} & PDDL-1.0 & 26 & 130 \\
\href{https://openneuro.org/datasets/ds000253}{ds000253} & Female action video game players. \citep{ds000253} & PDDL-1.0 & 20 & 100 \\
\href{https://openneuro.org/datasets/ds000254}{ds000254} & Multiband Multi-Echo Simultaneous ASL/BOLD for task-induced functional MRI \citep{ds000254} & CC0-1.0 & 13 & 13 \\
\href{https://openneuro.org/datasets/ds000256}{ds000256} & Behavioral interventions for reducing head motion during MRI scans in children \citep{ds000256} & PDDL-1.0 & 24 & 168 \\
\href{https://openneuro.org/datasets/ds000258}{ds000258} & Multi-echo Cambridge \citep{ds000258} & CC0-1.0 & 34 & 34 \\
\href{https://openneuro.org/datasets/ds001021}{ds001021} & NKI-sample \citep{ds001021} & not stated & 1 & 6 \\
\href{https://openneuro.org/datasets/ds001131}{ds001131} & Milky-Vodka \citep{ds001131} & CC0-1.0 & 54 & 72 \\
\href{https://openneuro.org/datasets/ds001132}{ds001132} & Sherlock \citep{ds001132} & not stated & 16 & 48 \\
\href{https://openneuro.org/datasets/ds001145}{ds001145} & Twilight Zone Movie Watching Dataset \citep{ds001145} & not stated & 24 & 24 \\
\href{https://openneuro.org/datasets/ds001226}{ds001226} & BTC\_preop \citep{ds001226} & CC0-1.0 & 36 & 36 \\
\href{https://openneuro.org/datasets/ds001229}{ds001229} & Neural Evidence of the Strategic Choice Between Working Memory and Episodic Memory in Prospective Remembering. \citep{ds001229} & not stated & 27 & 160 \\
\href{https://openneuro.org/datasets/ds001232}{ds001232} & Competition between items in working memory leads to forgetting \citep{ds001232} & not stated & 21 & 210 \\
\href{https://openneuro.org/datasets/ds001235}{ds001235} & Human-Object Interaction \citep{ds001235} & not stated & 12 & 168 \\
\href{https://openneuro.org/datasets/ds001241}{ds001241} & OLVSL\_ Object-location visual statistical learning \citep{ds001241} & CC0-1.0 & 24 & 234 \\
\href{https://openneuro.org/datasets/ds001242}{ds001242} & Examining effects of arousal on responses to salient and non-salient stimuli in younger and older adults \citep{ds001242} & PDDL-1.0 & 52 & 360 \\
\href{https://openneuro.org/datasets/ds001247}{ds001247} & sports\_betting\_study \citep{ds001247} & not stated & 43 & 43 \\
\href{https://openneuro.org/datasets/ds001297}{ds001297} & Personally familiar faces \citep{ds001297} & PDDL-1.0 & 33 & 363 \\
\href{https://openneuro.org/datasets/ds001302}{ds001302} & Inverse Base Rate \citep{ds001302} & PDDL-1.0 & 23 & 184 \\
\href{https://openneuro.org/datasets/ds001338}{ds001338} & ParanoiaStory \citep{ds001338} & CC-BY-NC-4.0 & 22 & 66 \\
\href{https://openneuro.org/datasets/ds001344}{ds001344} & phfMRI\_Control\_Tasks \citep{ds001344} & CC0-1.0 & 15 & 60 \\
\href{https://openneuro.org/datasets/ds001345}{ds001345} & Decoding of multisensory semantics and memories in low-level visual cortex \citep{ds001345} & CC0-1.0 & 12 & 48 \\
\href{https://openneuro.org/datasets/ds001353}{ds001353} & Information Filtering in Electronic Networks of Practice: An fMRI Investigation of Expectation [Dis]confirmation \citep{ds001353} & CC0-1.0 & 29 & 58 \\
\href{https://openneuro.org/datasets/ds001357}{ds001357} & NameFacePriming \citep{ds001357} & CC0-1.0 & 18 & 72 \\
\href{https://openneuro.org/datasets/ds001379}{ds001379} & Loci case study \citep{ds001379} & CC0-1.0 & 1 & 28 \\
\href{https://openneuro.org/datasets/ds001399}{ds001399} & multibandACQtests \citep{ds001399} & CC0-1.0 & 6 & 36 \\
\href{https://openneuro.org/datasets/ds001408}{ds001408} & rsfMRI\_single\_session\_EEG\_NF \citep{ds001408} & CC0-1.0 & 52 & 104 \\
\href{https://openneuro.org/datasets/ds001419}{ds001419} & thoughtExperiment \citep{ds001419} & CC0-1.0 & 1 & 5 \\
\href{https://openneuro.org/datasets/ds001430}{ds001430} & Neural Overlap in Item Representations Across Episodes Impairs Context Memory \citep{ds001430} & CC0-1.0 & 31 & 310 \\
\href{https://openneuro.org/datasets/ds001439}{ds001439} & A role for the medial temporal lobe subsystem in guiding prosociality: the effect of episodic processes on willingness to help others \citep{ds001439} & CC-BY-SA-4.0 & 18 & 179 \\
\href{https://openneuro.org/datasets/ds001454}{ds001454} & Human hippocampal replay during rest prioritizes weakly learned information and predicts memory performance \citep{ds001454} & CC0-1.0 & 24 & 432 \\
\href{https://openneuro.org/datasets/ds001491}{ds001491} & Valence processing differs across stimulus modalities (Multi-echo) \citep{ds001491} & CC0-1.0 & 20 & 95 \\
\href{https://openneuro.org/datasets/ds001521}{ds001521} & Lingering representations of stimuli influence recall organization \citep{ds001521} & not stated & 41 & 246 \\
\href{https://openneuro.org/datasets/ds001534}{ds001534} & Calorie-labeled food cues \citep{ds001534} & CC0-1.0 & 42 & 168 \\
\href{https://openneuro.org/datasets/ds001545}{ds001545} & Learning Naturalistic Temporal Structure in the Posterior Medial Network \citep{ds001545} & not stated & 30 & 90 \\
\href{https://openneuro.org/datasets/ds001553}{ds001553} & 100 runs at 3T \citep{ds001553} & CC0-1.0 & 3 & 351 \\
\href{https://openneuro.org/datasets/ds001554}{ds001554} & Neural pattern change during encoding of a narrative predicts retrospective duration estimates \citep{ds001554} & not stated & 23 & 23 \\
\href{https://openneuro.org/datasets/ds001566}{ds001566} & test \citep{ds001566} & not stated & 1 & 1 \\
\href{https://openneuro.org/datasets/ds001576}{ds001576} & Refresh my memory: Episodic memory reinstatements intrude on working memory maintenance \citep{ds001576} & not stated & 36 & 217 \\
\href{https://openneuro.org/datasets/ds001607}{ds001607} & Reinstated episodic context guides sampling-based decisions for reward \citep{ds001607} & not stated & 32 & 96 \\
\href{https://openneuro.org/datasets/ds001608}{ds001608} & Prestimulus Exp2 \citep{ds001608} & CC0-1.0 & 21 & 43 \\
\href{https://openneuro.org/datasets/ds001612}{ds001612} & Offline replay supports planning in human reinforcement learning \citep{ds001612} & not stated & 24 & 140 \\
\href{https://openneuro.org/datasets/ds001621}{ds001621} & Neural representations of events arise from temporal community structure \citep{ds001621} & not stated & 20 & 88 \\
\href{https://openneuro.org/datasets/ds001634}{ds001634} & Visual working memory: Study one Task fMRI and Behavioural response \citep{ds001634} & CC-BY-4.0 & 19 & 57 \\
\href{https://openneuro.org/datasets/ds001635}{ds001635} & Visual working memory: Study two Task fMRI and Behavioural response \citep{ds001635} & CC-BY-4.0 & 16 & 48 \\
\href{https://openneuro.org/datasets/ds001715}{ds001715} & Dissociable neural mechanisms track evidence accumulation for selection of attention versus action \citep{ds001715} & not stated & 34 & 408 \\
\href{https://openneuro.org/datasets/ds001722}{ds001722} & A role for the medial temporal lobe subsystem in guiding prosociality: the effect of episodic processes on willingness to help others (Experiment 2) \citep{ds001722} & CC-BY-SA-4.0 & 18 & 32 \\
\href{https://openneuro.org/datasets/ds001728}{ds001728} & Test-GEIER \citep{ds001728} & not stated & 1 & 1 \\
\href{https://openneuro.org/datasets/ds001740}{ds001740} & Brain activity during reciprocal social interaction investigated using conversational robots as control condition \citep{ds001740} & not stated & 25 & 100 \\
\href{https://openneuro.org/datasets/ds001745}{ds001745} & A neural signature of contextually mediated intentional forgetting \citep{ds001745} & not stated & 24 & 264 \\
\href{https://openneuro.org/datasets/ds001747}{ds001747} & Exploring the Resting State Neural Activity of Monolinguals and Late and Early Bilinguals \citep{ds001747} & CC0-1.0 & 92 & 92 \\
\href{https://openneuro.org/datasets/ds001748}{ds001748} & Differentiation of functional networks during long-term memory retrieval in children and adolescents \citep{ds001748} & CC-BY-SA-4.0 & 62 & 186 \\
\href{https://openneuro.org/datasets/ds001762}{ds001762} & Dorsal anterior cingulate and ventromedial prefrontal cortex have inverse roles in both foraging and economic choice \citep{ds001762} & not stated & 31 & 93 \\
\href{https://openneuro.org/datasets/ds001775}{ds001775} & Neural Differentiation Tracks Improved Recall of Competing Memories Following Interleaved Study and Retrieval Practice \citep{ds001775} & not stated & 24 & 144 \\
\href{https://openneuro.org/datasets/ds001796}{ds001796} & Bilingualism and the brain \citep{ds001796} & CC0-1.0 & 64 & 128 \\
\href{https://openneuro.org/datasets/ds001838}{ds001838} & Handedness and Symbolic Number Representation \citep{ds001838} & CC0-1.0 & 53 & 211 \\
\href{https://openneuro.org/datasets/ds001840}{ds001840} & Prestimulus Exp1 \citep{ds001840} & CC0-1.0 & 24 & 96 \\
\href{https://openneuro.org/datasets/ds001847}{ds001847} & A Probability Distribution over Latent Causes, in the Orbitofrontal Cortex \citep{ds001847} & not stated & 3 & 12 \\
\href{https://openneuro.org/datasets/ds001907}{ds001907} & ANT: Healthy aging and Parkinson's disease \citep{ds001907} & CC0-1.0 & 9 & 126 \\
\href{https://openneuro.org/datasets/ds001912}{ds001912} & S-cone\_MotionInDepth\_fMRI \citep{ds001912} & not stated & 15 & 137 \\
\href{https://openneuro.org/datasets/ds001921}{ds001921} & Anterior cingulate engagement in a foraging context reflects choice difficulty, not foraging value \citep{ds001921} & not stated & 15 & 30 \\
\href{https://openneuro.org/datasets/ds001923}{ds001923} & Anterior cingulate engagement in a foraging context reflects choice difficulty, not foraging value (2) \citep{ds001923} & not stated & 14 & 42 \\
\href{https://openneuro.org/datasets/ds001927}{ds001927} & The physiological effects of non-invasive brain stimulation fundamentally differ across the human cortex \citep{ds001927} & CC0-1.0 & 23 & 138 \\
\href{https://openneuro.org/datasets/ds001928}{ds001928} & Functional Connectivity of Music-Induced Analgesia in Fibromyalgia \citep{ds001928} & CC0-1.0 & 40 & 160 \\
\href{https://openneuro.org/datasets/ds002000}{ds002000} & reinstatement\_fidelity \citep{ds002000} & PDDL-1.0 & 21 & 168 \\
\href{https://openneuro.org/datasets/ds002011}{ds002011} & Locus coeruleus activity strengthens prioritized memories under arousal \citep{ds002011} & other (``PD'') & 22 & 110 \\
\href{https://openneuro.org/datasets/ds002013}{ds002013} & StimDisc \citep{ds002013} & CC0-1.0 & 4 & 36 \\
\href{https://openneuro.org/datasets/ds002014}{ds002014} & Language Production fMRI \citep{ds002014} & CC0-1.0 & 1 & 1 \\
\href{https://openneuro.org/datasets/ds002041}{ds002041} & Mesolimbic dopamine D2 receptors and neural representations of subjective value \citep{ds002041} & CC0-1.0 & 25 & 50 \\
\href{https://openneuro.org/datasets/ds002080}{ds002080} & BTC\_postop \citep{ds002080} & CC0-1.0 & 29 & 29 \\
\href{https://openneuro.org/datasets/ds002149}{ds002149} & Conditional Visual Associative Learning Task \citep{ds002149} & CC0-1.0 & 20 & 120 \\
\href{https://openneuro.org/datasets/ds002232}{ds002232} & Caltech rsfMRI Dataset \citep{ds002232} & CC0-1.0 & 6 & 12 \\
\href{https://openneuro.org/datasets/ds002250}{ds002250} & When my wrongs are worse than yours: behavioral and neural asymmetries in first-person and third-person perspectives of accidents \citep{ds002250} & CC0-1.0 & 20 & 223 \\
\href{https://openneuro.org/datasets/ds002270}{ds002270} & Viewing\_food\_cues \citep{ds002270} & CC0-1.0 & 19 & 38 \\
\href{https://openneuro.org/datasets/ds002322}{ds002322} & The Alice Dataset: fMRI Dataset to Study Natural Language Comprehension in the Brain \citep{ds002322} & CC0-1.0 & 26 & 26 \\
\href{https://openneuro.org/datasets/ds002345}{ds002345} & Narratives \citep{ds002345} & CC0-1.0 & 163 & 214 \\
\href{https://openneuro.org/datasets/ds002363}{ds002363} & Repetition of Computer Security Warnings Results in Differential Repetition Suppression Effects as Revealed with Functional MRI \citep{ds002363} & CC0-1.0 & 22 & 44 \\
\href{https://openneuro.org/datasets/ds002366}{ds002366} & Emotion regulation in the Ageing Brain, University of Reading, BBSRC \citep{ds002366} & CC0-1.0 & 34 & 136 \\
\href{https://openneuro.org/datasets/ds002522}{ds002522} & Contrast Response Functions \citep{ds002522} & CC0-1.0 & 48 & 188 \\
\href{https://openneuro.org/datasets/ds002547}{ds002547} & SharedStates \citep{ds002547} & CC0-1.0 & 26 & 115 \\
\href{https://openneuro.org/datasets/ds002603}{ds002603} & MB-EPI-comparison \citep{ds002603} & CC0-1.0 & 23 & 92 \\
\href{https://openneuro.org/datasets/ds002608}{ds002608} & RPN-signature\_Study1 \citep{ds002608} & CC0-1.0 & 41 & 41 \\
\href{https://openneuro.org/datasets/ds002609}{ds002609} & RPN-signature\_Study2 \citep{ds002609} & CC0-1.0 & 49 & 49 \\
\href{https://openneuro.org/datasets/ds002643}{ds002643} & Can we have a second helping? A replication study on the neurobiological mechanisms underlying self-control \citep{ds002643} & CC0-1.0 & 80 & 320 \\
\href{https://openneuro.org/datasets/ds002675}{ds002675} & The language network reemerges during recovery from severe traumatic brain injury \citep{ds002675} & CC0-1.0 & 24 & 64 \\
\href{https://openneuro.org/datasets/ds002711}{ds002711} & Smiley \citep{ds002711} & CC0-1.0 & 15 & 15 \\
\href{https://openneuro.org/datasets/ds002715}{ds002715} & AVScenes\_Blind \citep{ds002715} & CC0-1.0 & 8 & 31 \\
\href{https://openneuro.org/datasets/ds002717}{ds002717} & Unexplained Repeated Pregnancy Loss is Associated with Altered Perceptual and Brain Responses to Men's Body-Odor \citep{ds002717} & CC0-1.0 & 45 & 94 \\
\href{https://openneuro.org/datasets/ds002727}{ds002727} & EMOTIVE IDIOMS \citep{ds002727} & CC0-1.0 & 23 & 69 \\
\href{https://openneuro.org/datasets/ds002731}{ds002731} & Multiple interactive memory representations underlie the induction of false memory \citep{ds002731} & CC0-1.0 & 59 & 354 \\
\href{https://openneuro.org/datasets/ds002732}{ds002732} & Unnamed Dataset \citep{ds002732} & CC0-1.0 & 30 & 90 \\
\href{https://openneuro.org/datasets/ds002734}{ds002734} & Evidence Accumulation in Value-Based decisions \citep{ds002734} & CC0-1.0 & 22 & 44 \\
\href{https://openneuro.org/datasets/ds002737}{ds002737} & multibandCFtests \citep{ds002737} & CC0-1.0 & 1 & 9 \\
\href{https://openneuro.org/datasets/ds002748}{ds002748} & Resting state with closed eyes for patients with depression and healthy participants \citep{ds002748} & CC0-1.0 & 72 & 72 \\
\href{https://openneuro.org/datasets/ds002750}{ds002750} & YanDataBIDS \citep{ds002750} & CC0-1.0 & 3 & 3 \\
\href{https://openneuro.org/datasets/ds002785}{ds002785} & AOMIC-PIOP1 \citep{ds002785} & CC0-1.0 & 210 & 210 \\
\href{https://openneuro.org/datasets/ds002790}{ds002790} & AOMIC-PIOP2 \citep{ds002790} & CC0-1.0 & 226 & 896 \\
\href{https://openneuro.org/datasets/ds002793}{ds002793} & ToM following strong vs weak priors \citep{ds002793} & CC0-1.0 & 28 & 331 \\
\href{https://openneuro.org/datasets/ds002797}{ds002797} & eLife39497 \citep{ds002797} & CC0-1.0 & 22 & 198 \\
\href{https://openneuro.org/datasets/ds002835}{ds002835} & Prospection \citep{ds002835} & CC0-1.0 & 26 & 104 \\
\href{https://openneuro.org/datasets/ds002837}{ds002837} & Naturalistic Neuroimaging Database \citep{ds002837} & CC0-1.0 & 53 & 118 \\
\href{https://openneuro.org/datasets/ds002841}{ds002841} & Intuitive physics with fMRI \citep{ds002841} & CC0-1.0 & 29 & 174 \\
\href{https://openneuro.org/datasets/ds002842}{ds002842} & L2\_SENTENCES \citep{ds002842} & CC0-1.0 & 22 & 44 \\
\href{https://openneuro.org/datasets/ds002843}{ds002843} & Cognitive Training \citep{ds002843} & CC0-1.0 & 166 & 2{,}629 \\
\href{https://openneuro.org/datasets/ds002848}{ds002848} & mriquality \citep{ds002848} & CC0-1.0 & 5 & 5 \\
\href{https://openneuro.org/datasets/ds002872}{ds002872} & Aging \citep{ds002872} & CC0-1.0 & 39 & 196 \\
\href{https://openneuro.org/datasets/ds002896}{ds002896} & Resting State fMRI study of non bothersome tinnitus \citep{ds002896} & CC0-1.0 & 37 & 37 \\
\href{https://openneuro.org/datasets/ds002938}{ds002938} & Task-Dependent Effort-Induced Connectivity \citep{ds002938} & CC0-1.0 & 30 & 60 \\
\href{https://openneuro.org/datasets/ds002940}{ds002940} & Head-to-Ball Impacts in Collegiate Soccer Players \citep{ds002940} & CC0-1.0 & 24 & 47 \\
\href{https://openneuro.org/datasets/ds003007}{ds003007} & Two sessions of resting state with closed eyes for patients with depression in treatment course (NFB, CBT or No treatment groups) \citep{ds003007} & CC0-1.0 & 29 & 58 \\
\href{https://openneuro.org/datasets/ds003011}{ds003011} & Social Processes Initiative in Neurobiology of the Schizophrenia(s) Traveling Human Phantoms \citep{ds003011} & CC0-1.0 & 4 & 30 \\
\href{https://openneuro.org/datasets/ds003012}{ds003012} & action-in-auctions \citep{ds003012} & CC0-1.0 & 28 & 28 \\
\href{https://openneuro.org/datasets/ds003037}{ds003037} & SUDMEX\_TMS \citep{ds003037} & CC0-1.0 & 53 & 153 \\
\href{https://openneuro.org/datasets/ds003076}{ds003076} & Brain Correlates of Deductive Reasoning in Adults \citep{ds003076} & CC0-1.0 & 17 & 67 \\
\href{https://openneuro.org/datasets/ds003083}{ds003083} & Brain Correlates of Math Processing in Adults \citep{ds003083} & CC0-1.0 & 26 & 182 \\
\href{https://openneuro.org/datasets/ds003085}{ds003085} & Temporal Dynamics of Emotional Music \citep{ds003085} & CC0-1.0 & 39 & 156 \\
\href{https://openneuro.org/datasets/ds003089}{ds003089} & Somatosensory phase-encoded bilateral full-body light touch stimulation \citep{ds003089} & CC0-1.0 & 20 & 40 \\
\href{https://openneuro.org/datasets/ds003096}{ds003096} & Valuation of peers' safe choices is associated with substance-naivete in adolescents' \citep{ds003096} & CC0-1.0 & 31 & 31 \\
\href{https://openneuro.org/datasets/ds003103}{ds003103} & Consensus-seeking and conflict-resolving: an fMRI study on college couples' shopping interaction \citep{ds003103} & CC0-1.0 & 30 & 126 \\
\href{https://openneuro.org/datasets/ds003114}{ds003114} & How ovarian hormones influence the behaviroal activation and inhibition system through the dopamine pathway \citep{ds003114} & CC0-1.0 & 49 & 49 \\
\href{https://openneuro.org/datasets/ds003126}{ds003126} & Reading-related functional activity in children with isolated spelling deficits and dyslexia \citep{ds003126} & CC0-1.0 & 58 & 174 \\
\href{https://openneuro.org/datasets/ds003146}{ds003146} & vocal task \citep{ds003146} & CC0-1.0 & 33 & 33 \\
\href{https://openneuro.org/datasets/ds003171}{ds003171} & Modeling an auditory stimulated brain under altered states of consciousness using the generalized ising model \citep{ds003171} & CC0-1.0 & 17 & 136 \\
\href{https://openneuro.org/datasets/ds003342}{ds003342} & Hand-selective visual regions represent how to grasp 3D tools for use: brain decoding during real actions \citep{ds003342} & CC0-1.0 & 19 & 198 \\
\href{https://openneuro.org/datasets/ds003346}{ds003346} & SUDMEX\_CONN: The Mexican dataset of cocaine use disorder patients. \citep{ds003346} & CC0-1.0 & 141 & 141 \\
\href{https://openneuro.org/datasets/ds003354}{ds003354} & EmpaToM with PPI-R, TAS, and EQ \citep{ds003354} & CC0-1.0 & 10 & 10 \\
\href{https://openneuro.org/datasets/ds003381}{ds003381} & Estimation of in-scanner head pose changes during structural MRI using a convolutional neural network trained on eye tracker video \citep{ds003381} & CC0-1.0 & 21 & 21 \\
\href{https://openneuro.org/datasets/ds003382}{ds003382} & Monash vis-fPET-fMRI \citep{ds003382} & CC0-1.0 & 10 & 70 \\
\href{https://openneuro.org/datasets/ds003425}{ds003425} & Mental imagery can generate and regulate acquired differential fear conditioned reactivity \citep{ds003425} & CC0-1.0 & 13 & 317 \\
\href{https://openneuro.org/datasets/ds003427}{ds003427} & Effects of Phase Regression on High-Resolution Functional MRI of the Primary Visual Cortex \citep{ds003427} & CC0-1.0 & 4 & 24 \\
\href{https://openneuro.org/datasets/ds003436}{ds003436} & Agreeableness personality trait and social information encoding \citep{ds003436} & CC0-1.0 & 55 & 220 \\
\href{https://openneuro.org/datasets/ds003454}{ds003454} & Individual Differences in Fluid Reasoning and RAPM-like Problem Solving \citep{ds003454} & CC0-1.0 & 23 & 45 \\
\href{https://openneuro.org/datasets/ds003459}{ds003459} & Component processes of word reading in adults and children \citep{ds003459} & CC0-1.0 & 2 & 8 \\
\href{https://openneuro.org/datasets/ds003466}{ds003466} & Haptic three-dimensional curved surface exploration fMRI dataset \citep{ds003466} & CC0-1.0 & 21 & 135 \\
\href{https://openneuro.org/datasets/ds003469}{ds003469} & Speech disfluencies: Neurophysiological aspect in normal population \citep{ds003469} & CC0-1.0 & 81 & 81 \\
\href{https://openneuro.org/datasets/ds003487}{ds003487} & Differential contributions of ventral striatum subregions in the motivational and hedonic components of the affective response to reward \citep{ds003487} & CC0-1.0 & 24 & 48 \\
\href{https://openneuro.org/datasets/ds003495}{ds003495} & dStreamUpgrade \citep{ds003495} & CC0-1.0 & 24 & 138 \\
\href{https://openneuro.org/datasets/ds003500}{ds003500} & Response inhibition and selective attention in adults and children with and without ADHD \citep{ds003500} & CC0-1.0 & 38 & 223 \\
\href{https://openneuro.org/datasets/ds003507}{ds003507} & fMRI: Audiovisual Valence Congruence \citep{ds003507} & CC0-1.0 & 21 & 63 \\
\href{https://openneuro.org/datasets/ds003542}{ds003542} & Adult language learners \citep{ds003542} & CC0-1.0 & 34 & 320 \\
\href{https://openneuro.org/datasets/ds003545}{ds003545} & Adolescent language learners \citep{ds003545} & CC0-1.0 & 24 & 365 \\
\href{https://openneuro.org/datasets/ds003548}{ds003548} & Emotion Category and Face Perception Task Optimized for Multivariate Pattern Analysis \citep{ds003548} & CC0-1.0 & 16 & 96 \\
\href{https://openneuro.org/datasets/ds003550}{ds003550} & Visser, Scholte, Beemsterboer, \& Kindt (2013) Nature Neuroscience \citep{ds003550} & CC0-1.0 & 38 & 76 \\
\href{https://openneuro.org/datasets/ds003553}{ds003553} & Visser, Scholte \& Kindt (2011) Journal of Neuroscience \citep{ds003553} & CC0-1.0 & 18 & 18 \\
\href{https://openneuro.org/datasets/ds003554}{ds003554} & Visser et al. 2015 Psychoneuroendocrinology \citep{ds003554} & CC0-1.0 & 41 & 82 \\
\href{https://openneuro.org/datasets/ds003569}{ds003569} & Transposition confusability during visual word recognition \citep{ds003569} & CC0-1.0 & 24 & 24 \\
\href{https://openneuro.org/datasets/ds003574}{ds003574} & Reward biases spontaneous neural reactivation during sleep \citep{ds003574} & CC0-1.0 & 7 & 21 \\
\href{https://openneuro.org/datasets/ds003592}{ds003592} & Neurocognitive aging data release with behavioral, structural, and multi-echo functional MRI measures \citep{ds003592} & CC0-1.0 & 301 & 602 \\
\href{https://openneuro.org/datasets/ds003688}{ds003688} & Open multimodal iEEG-fMRI dataset from naturalistic stimulation with a short audiovisual film \citep{ds003688} & CC0-1.0 & 30 & 30 \\
\href{https://openneuro.org/datasets/ds003709}{ds003709} & NIMH-CompyPsych MMI \citep{ds003709} & CC0-1.0 & 51 & 153 \\
\href{https://openneuro.org/datasets/ds003763}{ds003763} & Interoception during aging: The heartbeat detection task \citep{ds003763} & CC0-1.0 & 29 & 29 \\
\href{https://openneuro.org/datasets/ds003770}{ds003770} & Depression Treatment by Medial Prefrontal Real-Time fMRI Neurofeedback \citep{ds003770} & CC0-1.0 & 7 & 95 \\
\href{https://openneuro.org/datasets/ds003772}{ds003772} & Functionally dissociable influences on learning rate in a dynamic environment \citep{ds003772} & CC0-1.0 & 32 & 128 \\
\href{https://openneuro.org/datasets/ds003789}{ds003789} & Dynamic changes in neural representations underlie the repetition effect on false memory \citep{ds003789} & CC0-1.0 & 32 & 384 \\
\href{https://openneuro.org/datasets/ds003831}{ds003831} & Cognitive Control Theoretic Mechanisms of Real-time fMRI-Guided Neuromodulation (CTM) \citep{ds003831} & CC0-1.0 & 70 & 346 \\
\href{https://openneuro.org/datasets/ds003835}{ds003835} & PE-Update \citep{ds003835} & CC0-1.0 & 24 & 165 \\
\href{https://openneuro.org/datasets/ds003849}{ds003849} & Individual differences in frontoparietal plasticity in humans \citep{ds003849} & CC0-1.0 & 91 & 181 \\
\href{https://openneuro.org/datasets/ds003858}{ds003858} & Monetary Incentive Delay task \citep{ds003858} & CC0-1.0 & 12 & 36 \\
\href{https://openneuro.org/datasets/ds003871}{ds003871} & Resting-state for 34 younger and 28 older adults \citep{ds003871} & CC0-1.0 & 62 & 62 \\
\href{https://openneuro.org/datasets/ds003892}{ds003892} & Loss and enhancement of layer-selective signals in geniculostriate and corticotectal pathways of adult human amblyopia \citep{ds003892} & CC0-1.0 & 34 & 270 \\
\href{https://openneuro.org/datasets/ds003965}{ds003965} & NI-edu-data-minimal \citep{ds003965} & CC0-1.0 & 1 & 14 \\
\href{https://openneuro.org/datasets/ds003972}{ds003972} & Body Motor \citep{ds003972} & CC0-1.0 & 8 & 30 \\
\href{https://openneuro.org/datasets/ds003990}{ds003990} & fastVsSlowERtactile7T \citep{ds003990} & CC0-1.0 & 6 & 130 \\
\href{https://openneuro.org/datasets/ds003999}{ds003999} & Pre-Post rehabilitation fMRI data of post-stroke patients. \citep{ds003999} & CC0-1.0 & 26 & 52 \\
\href{https://openneuro.org/datasets/ds004021}{ds004021} & Visuomotor Rotation Adaptation Experiment \citep{ds004021} & CC0-1.0 & 32 & 320 \\
\href{https://openneuro.org/datasets/ds004101}{ds004101} & Resting-state fMRI data acquired before and after a course of infra-low frequency neurofeedback \citep{ds004101} & CC0-1.0 & 9 & 18 \\
\href{https://openneuro.org/datasets/ds004128}{ds004128} & Neural representations of others' traits predict social decisions \citep{ds004128} & CC0-1.0 & 32 & 64 \\
\href{https://openneuro.org/datasets/ds004134}{ds004134} & Experience Sampling in Resting State \citep{ds004134} & CC0-1.0 & 36 & 72 \\
\href{https://openneuro.org/datasets/ds004141}{ds004141} & Targeting Dynamic Facial Processing Mechanisms in Superior Temporal Sulcus Using fMRI Neurofeedback \citep{ds004141} & CC0-1.0 & 20 & 80 \\
\href{https://openneuro.org/datasets/ds004142}{ds004142} & Exploring the Neural Correlates of Feedback-Related Reward Saliency and Valence During Real-Time fMRI-Based Neurofeedback \citep{ds004142} & CC0-1.0 & 9 & 45 \\
\href{https://openneuro.org/datasets/ds004144}{ds004144} & A behavioral, clinical and brain imaging dataset with focus on emotion regulation of females with fibromyalgia \citep{ds004144} & CC0-1.0 & 66 & 131 \\
\href{https://openneuro.org/datasets/ds004182}{ds004182} & Magic, Memory, and Curiosity (MMC) fMRI Dataset \citep{ds004182} & CC0-1.0 & 50 & 251 \\
\href{https://openneuro.org/datasets/ds004196}{ds004196} & Bimodal dataset on Inner speech \citep{ds004196} & CC0-1.0 & 4 & 8 \\
\href{https://openneuro.org/datasets/ds004213}{ds004213} & BrainBeats \citep{ds004213} & CC0-1.0 & 5 & 10 \\
\href{https://openneuro.org/datasets/ds004219}{ds004219} & Litebook\_Alertness\_Study \citep{ds004219} & CC0-1.0 & 10 & 20 \\
\href{https://openneuro.org/datasets/ds004302}{ds004302} & Brain correlates of speech perception in schizophrenia patients with and without auditory hallucinations \citep{ds004302} & CC0-1.0 & 71 & 71 \\
\href{https://openneuro.org/datasets/ds004392}{ds004392} & Parkinson's disease, functional connectivity, and cognition \citep{ds004392} & CC0-1.0 & 58 & 58 \\
\href{https://openneuro.org/datasets/ds004400}{ds004400} & Number comparison task \citep{ds004400} & CC0-1.0 & 25 & 200 \\
\href{https://openneuro.org/datasets/ds004440}{ds004440} & Stanford Child and Adult Checkerboard Retinotopy Dataset \citep{ds004440} & CC0-1.0 & 49 & 195 \\
\href{https://openneuro.org/datasets/ds004450}{ds004450} & XCP\_WALKTHROUGH \citep{ds004450} & CC0-1.0 & 3 & 6 \\
\href{https://openneuro.org/datasets/ds004467}{ds004467} & MRI data from 20 adults in response to videos of dialogue and monologue from Sesame Street \citep{ds004467} & CC0-1.0 & 18 & 177 \\
\href{https://openneuro.org/datasets/ds004469}{ds004469} & Temporal Lobe Epilepsy - UNAM \citep{ds004469} & CC0-1.0 & 66 & 128 \\
\href{https://openneuro.org/datasets/ds004498}{ds004498} & Perinatal Stroke \citep{ds004498} & CC0-1.0 & 1 & 120 \\
\href{https://openneuro.org/datasets/ds004544}{ds004544} & An fMRI Investigation of the Neurocognitive Processing of Strategies and Measures \citep{ds004544} & CC0-1.0 & 38 & 114 \\
\href{https://openneuro.org/datasets/ds004557}{ds004557} & The role of superstition of cognitive control during neurofeedback training - part 2 \citep{ds004557} & CC0-1.0 & 35 & 35 \\
\href{https://openneuro.org/datasets/ds004562}{ds004562} & MVPA for motor adaptation \citep{ds004562} & CC0-1.0 & 23 & 23 \\
\href{https://openneuro.org/datasets/ds004564}{ds004564} & Demo \citep{ds004564} & CC0-1.0 & 2 & 2 \\
\href{https://openneuro.org/datasets/ds004594}{ds004594} & Brain surface fingerprints \citep{ds004594} & CC0-1.0 & 20 & 80 \\
\href{https://openneuro.org/datasets/ds004604}{ds004604} & Cerebrovascular Reactivity Normative Dataset \citep{ds004604} & CC0-1.0 & 50 & 50 \\
\href{https://openneuro.org/datasets/ds004627}{ds004627} & Characterization and Treatment of Adolescent Depression (CAT-D) \citep{ds004627} & CC0-1.0 & 130 & 962 \\
\href{https://openneuro.org/datasets/ds004648}{ds004648} & Sympathovagal quotient and functional connectivity of control networks are related to gut Ruminococcaceae abundance in healthy men \citep{ds004648} & CC0-1.0 & 88 & 88 \\
\href{https://openneuro.org/datasets/ds004656}{ds004656} & Neural reactivity to visual food stimuli is reduced in some areas of the brain during evening hours compared to morning hours: an fMRI study in women \citep{ds004656} & CC0-1.0 & 15 & 60 \\
\href{https://openneuro.org/datasets/ds004662}{ds004662} & Denoising task-correlated head motion from motor-task fMRI data with multi-echo ICA \citep{ds004662} & CC0-1.0 & 8 & 16 \\
\href{https://openneuro.org/datasets/ds004697}{ds004697} & Food and Brain Study \citep{ds004697} & CC0-1.0 & 82 & 482 \\
\href{https://openneuro.org/datasets/ds004718}{ds004718} & Le Petit Prince Hong Kong: Naturalistic fMRI and EEG dataset from older Cantonese speakers \citep{ds004718} & CC0-1.0 & 50 & 250 \\
\href{https://openneuro.org/datasets/ds004720}{ds004720} & Descriptive name for this dataset \citep{ds004720} & CC0-1.0 & 2 & 8 \\
\href{https://openneuro.org/datasets/ds004725}{ds004725} & Single Dose Intranasal Oxytocin Administration: Data from Healthy Younger and Older Adults \citep{ds004725} & CC0-1.0 & 87 & 87 \\
\href{https://openneuro.org/datasets/ds004737}{ds004737} & CS-DSI \citep{ds004737} & CC0-1.0 & 20 & 20 \\
\href{https://openneuro.org/datasets/ds004765}{ds004765} & language fMRI \citep{ds004765} & CC0-1.0 & 71 & 71 \\
\href{https://openneuro.org/datasets/ds004775}{ds004775} & Vicarious Punishment of Moral Violations in Naturalistic Drama Narratives Predicts Cortical Synchronization \citep{ds004775} & CC0-1.0 & 28 & 84 \\
\href{https://openneuro.org/datasets/ds004783}{ds004783} & Functional MRI dataset for evaluating brain synchrony among dance spectators \citep{ds004783} & CC0-1.0 & 26 & 26 \\
\href{https://openneuro.org/datasets/ds004798}{ds004798} & fMRI data for: Multimodal single-neuron, intracranial EEG, and fMRI brain responses during movie watching in human patients \citep{ds004798} & CC0-1.0 & 15 & 30 \\
\href{https://openneuro.org/datasets/ds004835}{ds004835} & Processing of frequency and location in human subcortical auditory structures \citep{ds004835} & CC0-1.0 & 6 & 48 \\
\href{https://openneuro.org/datasets/ds004848}{ds004848} & Game of Thrones - A naturalistic viewing dataset \citep{ds004848} & CC0-1.0 & 73 & 145 \\
\href{https://openneuro.org/datasets/ds004873}{ds004873} & BOLD signal changes can oppose oxygen metabolism across the human cortex \citep{ds004873} & CC0-1.0 & 73 & 90 \\
\href{https://openneuro.org/datasets/ds004956}{ds004956} & Braille letters - fMRI \citep{ds004956} & CC0-1.0 & 15 & 266 \\
\href{https://openneuro.org/datasets/ds005016}{ds005016} & MBSR \citep{ds005016} & CC0-1.0 & 147 & 1{,}694 \\
\href{https://openneuro.org/datasets/ds005027}{ds005027} & Michigan Longitudinal Study \citep{ds005027} & CC0-1.0 & 81 & 324 \\
\href{https://openneuro.org/datasets/ds005040}{ds005040} & Political Moralization (PMM) \citep{ds005040} & CC0-1.0 & 50 & 441 \\
\href{https://openneuro.org/datasets/ds005069}{ds005069} & test \citep{ds005069} & CC0-1.0 & 1 & 4 \\
\href{https://openneuro.org/datasets/ds005072}{ds005072} & test \citep{ds005072} & CC0-1.0 & 1 & 4 \\
\href{https://openneuro.org/datasets/ds005075}{ds005075} & Simultaneous Brain and Spinal Cord at rest: an fMRI dataset \citep{ds005075} & CC0-1.0 & 30 & 30 \\
\href{https://openneuro.org/datasets/ds005128}{ds005128} & Deception Signaling Task \citep{ds005128} & CC0-1.0 & 40 & 80 \\
\href{https://openneuro.org/datasets/ds005134}{ds005134} & Sequence test for multiband DWI and magnetization transfer sequence \citep{ds005134} & CC0-1.0 & 1 & 1 \\
\href{https://openneuro.org/datasets/ds005166}{ds005166} & A framework for leveraging technology to probe mechanism: A proof of concept study of inhibitory control \citep{ds005166} & CC0-1.0 & 39 & 156 \\
\href{https://openneuro.org/datasets/ds005230}{ds005230} & Reinforcement-Learning Generalization \citep{ds005230} & CC0-1.0 & 44 & 130 \\
\href{https://openneuro.org/datasets/ds005239}{ds005239} & motor-fmri \citep{ds005239} & CC0-1.0 & 4 & 75 \\
\href{https://openneuro.org/datasets/ds005339}{ds005339} & Yale\_reading\_dataset \citep{ds005339} & CC0-1.0 & 217 & 1{,}828 \\
\href{https://openneuro.org/datasets/ds005355}{ds005355} & BSC\_BIDS\_071724 \citep{ds005355} & CC0-1.0 & 12 & 88 \\
\href{https://openneuro.org/datasets/ds005365}{ds005365} & Neurobiological substrates of Altered State of Consciousness Induced by High Ventilation Breathwork \citep{ds005365} & CC0-1.0 & 1 & 4 \\
\href{https://openneuro.org/datasets/ds005418}{ds005418} & Cerebrovascular Reactivity: CO2 inhalation and resting state comparisons \citep{ds005418} & CC0-1.0 & 35 & 70 \\
\href{https://openneuro.org/datasets/ds005454}{ds005454} & fMRIPrep boot camp -- 7T training dataset \citep{ds005454} & CC0-1.0 & 5 & 10 \\
\href{https://openneuro.org/datasets/ds005455}{ds005455} & An fMRI dataset for investigating language control and cognitive control in bilinguals \citep{ds005455} & CC0-1.0 & 77 & 299 \\
\href{https://openneuro.org/datasets/ds005479}{ds005479} & Monetary Incentive Delay task - structural and functional images of 37 men; study of associations between circadian characteristics (eveningness, distinctness) and affective processing \citep{ds005479} & CC0-1.0 & 37 & 37 \\
\href{https://openneuro.org/datasets/ds005492}{ds005492} & Neurowater Project \citep{ds005492} & CC0-1.0 & 35 & 35 \\
\href{https://openneuro.org/datasets/ds005504}{ds005504} & Psychosocial adversity and inhibitory control: an fMRI study of children growing up in extreme poverty \citep{ds005504} & CC0-1.0 & 68 & 204 \\
\href{https://openneuro.org/datasets/ds005518}{ds005518} & Deciphering the neural responses to a naturalistic persuasive message \citep{ds005518} & CC0-1.0 & 50 & 50 \\
\href{https://openneuro.org/datasets/ds005525}{ds005525} & Circadian misalignment and energy balance \citep{ds005525} & CC0-1.0 & 11 & 349 \\
\href{https://openneuro.org/datasets/ds005530}{ds005530} & Depotentiation of emotional reactivity using TMR during REM sleep \citep{ds005530} & CC0-1.0 & 18 & 18 \\
\href{https://openneuro.org/datasets/ds005533}{ds005533} & High-quality multimodal MRI with simultaneous EEG using conductive ink and polymer-thick film nets \citep{ds005533} & CC0-1.0 & 3 & 12 \\
\href{https://openneuro.org/datasets/ds005573}{ds005573} & A comparison of resting state functional magnetic resonance imaging to invasive electrocortical stimulation for sensorimotor mapping in pediatric patients \citep{ds005573} & CC0-1.0 & 16 & 32 \\
\href{https://openneuro.org/datasets/ds005588}{ds005588} & SoDec - Responsibility fMRI experiment dataset \citep{ds005588} & CC0-1.0 & 40 & 80 \\
\href{https://openneuro.org/datasets/ds005596}{ds005596} & A naturalistic fMRI dataset in response to puclic speaking \citep{ds005596} & CC0-1.0 & 12 & 20 \\
\href{https://openneuro.org/datasets/ds005598}{ds005598} & Visuomotor Rotation Learning and Reward-based motor learning \citep{ds005598} & CC0-1.0 & 45 & 308 \\
\href{https://openneuro.org/datasets/ds005603}{ds005603} & Mitchell\_Hacker\_2013 \citep{ds005603} & CC0-1.0 & 8 & 52 \\
\href{https://openneuro.org/datasets/ds005623}{ds005623} & test \citep{ds005623} & CC0-1.0 & 68 & 204 \\
\href{https://openneuro.org/datasets/ds005625}{ds005625} & Connectomix test dataset 1 \citep{ds005625} & CC0-1.0 & 1 & 1 \\
\href{https://openneuro.org/datasets/ds005699}{ds005699} & Connectomix test dataset 2 \citep{ds005699} & CC0-1.0 & 4 & 4 \\
\href{https://openneuro.org/datasets/ds005754}{ds005754} & Multivariate Assessment of Inhibitory Control in Youth: Links with Psychopathology and Brain Function Dataset \citep{ds005754} & CC0-1.0 & 124 & 124 \\
\href{https://openneuro.org/datasets/ds005783}{ds005783} & Representation of navigational affordances and ego-motion in the occipital place area - dataset \citep{ds005783} & CC0-1.0 & 16 & 172 \\
\href{https://openneuro.org/datasets/ds005795}{ds005795} & MULTI-CLARID (Multimodal Category Learning and Resting-state Imaging Data) \citep{ds005795} & CC0-1.0 & 34 & 39 \\
\href{https://openneuro.org/datasets/ds005850}{ds005850} & Spoken narrative comprehension in Chinese \citep{ds005850} & CC0-1.0 & 68 & 204 \\
\href{https://openneuro.org/datasets/ds005882}{ds005882} & Integration of contextual reward motivation with category representations \citep{ds005882} & CC0-1.0 & 23 & 71 \\
\href{https://openneuro.org/datasets/ds005883}{ds005883} & CoSpine Database\_pain\_dataset \citep{ds005883} & CC0-1.0 & 39 & 39 \\
\href{https://openneuro.org/datasets/ds005884}{ds005884} & CoSpine Database\_motor\_dataset \citep{ds005884} & CC0-1.0 & 22 & 44 \\
\href{https://openneuro.org/datasets/ds005892}{ds005892} & Resting State MRI data from healthy control (HC), Parkinson's disease with normal cognition (PD-NC), and Parkinson's disease with mild cognitive impairment (PD-MCI) cohorts \citep{ds005892} & CC0-1.0 & 55 & 55 \\
\href{https://openneuro.org/datasets/ds005903}{ds005903} & Evidence against a cortical module for processing communicative gaze \citep{ds005903} & CC0-1.0 & 6 & 120 \\
\href{https://openneuro.org/datasets/ds006092}{ds006092} & CogRief Study \citep{ds006092} & CC0-1.0 & 60 & 60 \\
\href{https://openneuro.org/datasets/ds006156}{ds006156} & Executive Functioning-Related Neural Processes in Emerging Adults with Type 1 Diabetes: A Functional Magnetic Resonance Imaging Study \citep{ds006156} & CC0-1.0 & 62 & 248 \\
\href{https://openneuro.org/datasets/ds006193}{ds006193} & PAFIN: PennLINC AFfective INstability (Test Dataset) \citep{ds006193} & CC0-1.0 & 1 & 1 \\
\href{https://openneuro.org/datasets/ds006211}{ds006211} & Brain bases for navigating acoustic features - fMRI dataset \citep{ds006211} & CC0-1.0 & 25 & 96 \\
\href{https://openneuro.org/datasets/ds006559}{ds006559} & Attitude Change Study 1 Persuade \citep{ds006559} & CC0-1.0 & 35 & 173 \\
\href{https://openneuro.org/datasets/ds006568}{ds006568} & Attitude Change Study 2 Debate \citep{ds006568} & CC0-1.0 & 37 & 222 \\
\href{https://openneuro.org/datasets/ds006572}{ds006572} & RSfMRI\_Vision \citep{ds006572} & CC0-1.0 & 20 & 200 \\
\href{https://openneuro.org/datasets/ds006592}{ds006592} & The First Comprehensive Study of Early-Emerging Prosopometamorphopsia \citep{ds006592} & CC0-1.0 & 1 & 4 \\
\href{https://openneuro.org/datasets/ds006594}{ds006594} & RSfMRI\_Function \citep{ds006594} & CC0-1.0 & 20 & 200 \\
\href{https://openneuro.org/datasets/ds006729}{ds006729} & R01\_Neuromuscular\_Signature\_Pilot \citep{ds006729} & CC0-1.0 & 28 & 112 \\
\href{https://openneuro.org/datasets/ds006893}{ds006893} & The interplay of central insulin and menstrual cycle on functional brain networks and neural food cue reactivity in women \citep{ds006893} & CC0-1.0 & 15 & 108 \\
\href{https://openneuro.org/datasets/ds006944}{ds006944} & One-person corticospinal fMRI dataset for Stanford Spinal Cord workshop 2025 \citep{ds006944} & CC0-1.0 & 1 & 2 \\
\href{https://openneuro.org/datasets/ds006945}{ds006945} & Dataset: T1-Weighted Structural MRI and fMRI of Participants Viewing Self-Avatar Exoskeleton Walking (11 SWS Cycles) \citep{ds006945} & CC0-1.0 & 5 & 14 \\
\href{https://openneuro.org/datasets/ds006952}{ds006952} & PSYCH-REST \citep{ds006952} & CC0-1.0 & 34 & 271 \\
\href{https://openneuro.org/datasets/ds007115}{ds007115} & Neural Mechanisms of Cognitive Conflict: Processing COVID-19 Vaccine Misinformation \citep{ds007115} & CC0-1.0 & 39 & 117 \\
\href{https://openneuro.org/datasets/ds007126}{ds007126} & An open fMRI dataset examining the effects of online social and non-social information distraction on global-local attention \citep{ds007126} & CC0-1.0 & 34 & 102 \\
\href{https://openneuro.org/datasets/ds007135}{ds007135} & Investigating Neuroepigenetic Alterations in Chronic Low Back Pain with PET/MR \citep{ds007135} & CC0-1.0 & 25 & 106 \\
\href{https://openneuro.org/datasets/ds007185}{ds007185} & Neural Responses to Curiosity in Young Children \citep{ds007185} & CC0-1.0 & 18 & 36 \\
\href{https://openneuro.org/datasets/ds007313}{ds007313} & Brain and spinal cord fMRI and qMRI - Single participant \citep{ds007313} & CC0-1.0 & 1 & 1 \\
\href{https://openneuro.org/datasets/ds007345}{ds007345} & Multimodal Brain MRI and Clinical Data in Olfactory Groove Meningioma: A Prospective Data Report \citep{ds007345} & CC0-1.0 & 56 & 56 \\
\href{https://openneuro.org/datasets/ds007386}{ds007386} & MOFOMIC-NIMF3 \citep{ds007386} & CC0-1.0 & 27 & 103 \\
\href{https://openneuro.org/datasets/ds007441}{ds007441} & Human Dignity \citep{ds007441} & CC0-1.0 & 40 & 40 \\
\href{https://openneuro.org/datasets/ds007534}{ds007534} & Neuroil \citep{ds007534} & CC0-1.0 & 34 & 34 \\
\midrule
\textbf{Total} & 322 datasets & & 11{,}379 & 45{,}731 \\
\end{longtable}
}